\documentclass[10pt,twocolumn,letterpaper]{article}

\usepackage[pagenumbers]{wacv} 

\usepackage{graphicx}
\usepackage{amsmath}
\usepackage{amssymb}
\usepackage{booktabs}

\usepackage{color}

\usepackage{multirow}
\usepackage{dashrule}
\usepackage{float}
\usepackage{subcaption}
\usepackage{placeins}

\usepackage[accsupp]{axessibility}

\definecolor{cvprblue}{rgb}{0.21,0.49,0.74}

\definecolor{textgreen}{rgb}{0.4980392156862745, 0.788235294117647, 0.4980392156862745}
\definecolor{ourmediumblue}{rgb}{0.21568627450980393,0.49411764705882355,0.7215686274509804}
\definecolor{ourmediumred}{rgb}{0.8941176470588236,0.10196078431372549,0.10980392156862745}
\definecolor{ourmediumgreen}{rgb}{0.30196078431372547,0.6862745098039216,0.2901960784313726}

\definecolor{ourlightred}{rgb}{0.984313725490196, 0.5019607843137255, 0.4470588235294118}
\definecolor{ourlightblue}{rgb}{0.5019607843137255, 0.6941176470588235, 0.8274509803921568}
\definecolor{ourlightcyan}{rgb}{0.5529411764705883, 0.8274509803921568, 0.7803921568627451}

\definecolor{americanrose}{rgb}{1.0, 0.01, 0.24}
\definecolor{amethyst}{rgb}{0.6, 0.4, 0.8}

\newcommand{\txtgreen}[1]{\textcolor{textgreen}{#1}}

\usepackage[dvipsnames]{xcolor}

\newcommand{\pointToAppendix}[2]{Supp. #2}

\newcommand\extrafootertext[1]{%
    \bgroup
    \renewcommand\thefootnote{\fnsymbol{footnote}}%
    \renewcommand\thempfootnote{\fnsymbol{mpfootnote}}%
    \footnotetext[0]{#1}%
    \egroup
}

\usepackage[pagebackref,breaklinks,colorlinks]{hyperref}

\usepackage[capitalize]{cleveref}
\crefname{section}{Sec.}{Secs.}
\Crefname{section}{Section}{Sections}
\Crefname{table}{Table}{Tables}
\crefname{table}{Tab.}{Tabs.}

\def\wacvPaperID{309} 
\def\confName{WACV}
\def\confYear{2025}

\begin{document}

\title{Instance-Warp: Saliency Guided Image Warping for Unsupervised Domain Adaptation}


\author{
Shen Zheng\thanks{Equal contribution} \quad\quad Anurag Ghosh\footnotemark[1] \quad\quad Srinivasa G. Narasimhan \\
Carnegie Mellon University \\
{\tt\small \{shenzhen, anuraggh, srinivas\}@cs.cmu.edu} \\
}




\maketitle









\begin{abstract}
\vspace{-0.15in}

Driving is challenging in conditions like night, rain, and snow. Lack of good labeled datasets has hampered progress in scene understanding under such conditions. Unsupervised Domain Adaptation (UDA) using large labeled clear-day datasets is a promising research direction in such cases. However, many UDA methods are trained with dominant scene backgrounds (e.g., roads, sky, sidewalks) that appear dramatically different across domains. As a result, they struggle to learn effective features of smaller and often sparse foreground objects (e.g., people, vehicles, signs).



In this work, we improve UDA training by applying in-place image warping to focus on salient objects. We design instance-level saliency guidance to adaptively oversample object regions and undersample background areas, which reduces adverse effects from background context and enhances backbone feature learning. Our approach improves adaptation across geographies, lighting, and weather conditions, and is agnostic to the task (segmentation, detection), domain adaptation algorithm, saliency guidance, and underlying model architecture. Result highlights include \textbf{\txtgreen{+6.1}} mAP50 for BDD100K Clear $\rightarrow$ DENSE Foggy, \textbf{\txtgreen{+3.7}} mAP50 for BDD100K Day $\rightarrow$ Night, \textbf{\txtgreen{+3.0}} mAP50 for BDD100K Clear $\rightarrow$ Rainy, and \textbf{\txtgreen{+6.3}}  mIoU for Cityscapes $\rightarrow$ ACDC. 
Besides, Our method adds minimal training memory and no additional inference latency. Code is available at \href{https://github.com/ShenZheng2000/Instance-Warp}{https://github.com/ShenZheng2000/Instance-Warp}.


\end{abstract}    
\vspace{-0.15in}
\section{Introduction}
\label{sec:intro}



\begin{figure}[t!]
    \centering
    \includegraphics[width=\linewidth]{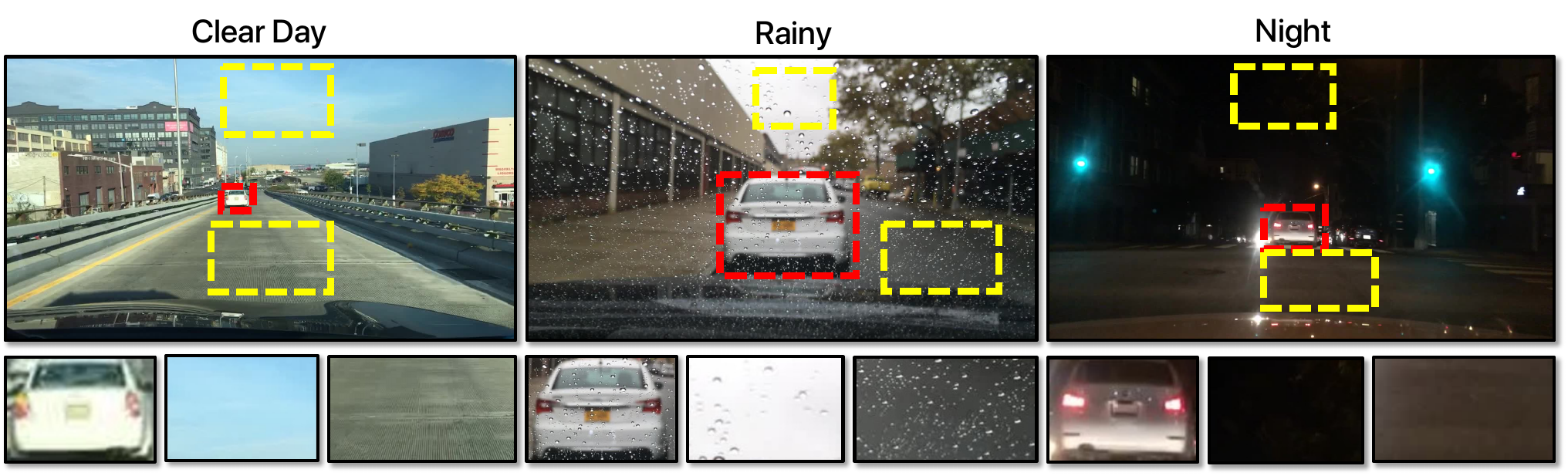}
    \vspace{-0.1in}
    \caption{\textbf{Object-Background Pixel Imbalance \& Differences in Cross-Domain Object-Background Variations.} Consider scenes from different domains: clear day, rainy, and night. \colorbox{Red}{\textbf{\textcolor{black}{red}}} highlights foreground objects and \colorbox{yellow}{\textbf{\textcolor{black}{yellow}}} highlights background areas. Note, \textit{(a)} background pixels occupy more space than foreground object pixels, \textit{(b)} background elements like road exhibit higher cross-domain variations, while foreground objects like cars show smaller variations. Thus, focusing on objects would mitigate over-reliance on the background and improve domain adaptation.}
    \label{fig:method-overview}
    \vspace{-0.25in}
\end{figure}

\noindent
Consider driving in safety-critical scenarios such as at night, in rain, or snow. Robust detection and segmentation of objects (e.g., cars and pedestrians) and background are critical to prevent accidents. In these conditions, human perception focuses on salient objects rather than background for decision making~\cite{underwood2011decisions}, using high-level semantics to search and decide where to focus~\cite{hayes2019scene}. 

Camera-captured images, meanwhile, exhibit an \textit{object-background pixel imbalance}, where pixel distribution is skewed towards backgrounds taking up more space. In contrast, objects are smaller (e.g. traffic signs), more scattered, occupying fewer pixels. The camera's perspective further amplifies this effect, distant objects appear smaller and larger backgrounds to dominate the scene. However, many contemporary perception models treat image pixels, whether objects or backgrounds, uniformly. This leads to an over-reliance on the dominant backgrounds, making it challenging to learn effective features for less prevalent objects amid these dominant backgrounds.



\begin{figure}[t!]
    \centering
    \includegraphics[width=\linewidth]{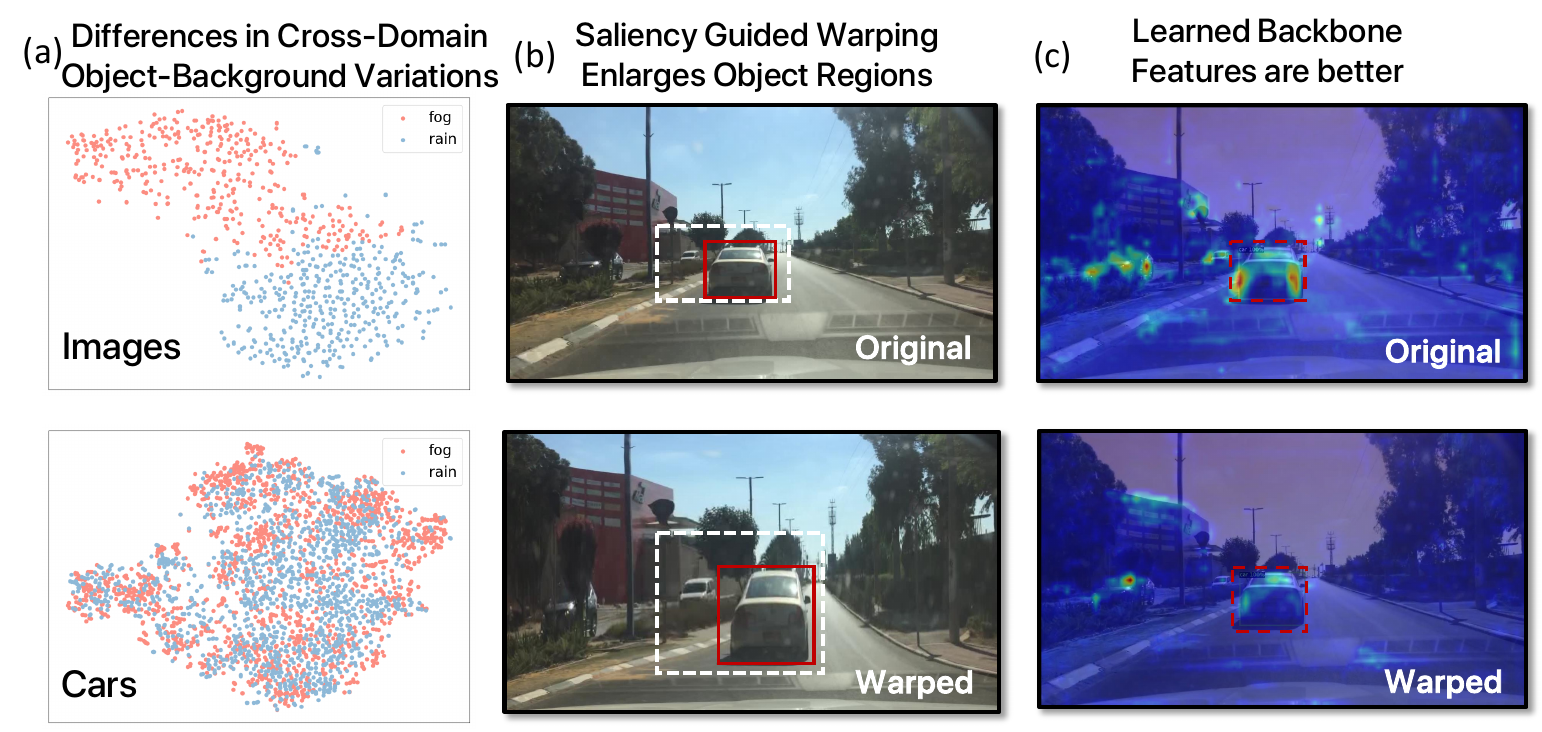}
    \vspace{-0.2in}
    \caption{\textit{(a)} \textbf{Differences in Cross-domain object background variations.} We visualize t-SNE~\cite{van2008visualizing} plots for ResNet-50 features extracted from ACDC~\cite{sakaridis2021acdc} foggy and rainy images. Top plot shows image features (the majority of an image is background), while the bottom plot shows car features (the most common object). While image features across domains exhibit higher variance, car features are tightly entangled. Thus, over-reliance on the more variable background context makes adaptation difficult. \textit{(b)} \textbf{Saliency Guided Image Warping increases the relative size of salient foreground regions.} Zoom in to see the car getting enlarged in the highlighted region, which reduces the effect of background context. \textit{(c)} \textbf{Learned Backbone features are better} - they show a better focus on the object and less reliance on background.}
    \label{fig:method-motivation}
    \vspace{-0.2in}
\end{figure}

While this over-reliance on background pixels is a significant issue in traditional vision tasks, in Unsupervised Domain Adaptation (UDA)~\cite{hoyer2022daformer, hoyer2022hrda, kennerley20232pcnet}, which is crucial for driving in challenging conditions due to the scarcity of labeled data, the problem is further compounded by \textit{differences in cross-domain object-background variations}, where background regions show greater variations than object regions across domains (Figure~\ref{fig:method-motivation}). Consequently, this over-reliance of UDA algorithms on background pixels makes adaptation difficult. Reducing dependence on these variable backgrounds would improve adaptation to new domains.

One possible remedy is data augmentation (e.g., resizing \textit{viz} multi-scale training~\cite{singh2018analysis, hoyer2022hrda} and inference~\cite{he2016deep, shrivastava2016training, hoyer2022hrda}). However, these strategies introduce significant computational and storage overhead. Another approach, employing architectural priors to address these issues (e.g., feature pyramids~\cite{lin2017feature, lowe2004distinctive, liu2018path, li2020spatial}, coarse-to-fine feature alignment~\cite{zheng2020cross} or test-time scale optimization~\cite{ghosh2023chanakya, chin2019adascale}), only partially improve recognition with additional inference latency. Yet another strategy is to oversample or `zoom' into salient image regions~\cite{recasens2018learning, thavamani2021fovea, ehteshami2022salisa, ghosh2023learned, thavamani2023learning}. However, these methods do not distinguish between object and background, are only applicable to supervised settings, and introduce additional inference latency due to test-time operations. 


To address these limitations, our novel strategy judiciously utilizes available pixel space in an image to focus on salient regions for unsupervised domain adaptation. We \textbf{adaptively oversample salient image regions corresponding to objects in-place}, directly warping the original image without creating new images or additional crops. After that, we \textbf{unwarp improved features while predicting labels} and adapt models from a source to a target domain. This reduces background context dependence (Figure~\ref{fig:method-motivation}), significantly improving domain adaptability. During inference, we do not perform any warping as the learned features are already improved, thus adding no latency. We show significant improvements on real to real adaptation scenarios~\cite{hoyer2022daformer, kennerley20232pcnet, hoyer2022hrda, hoyer2023mic} (Sections~\ref{subsec:daod-results},~\ref{subsec:dass-results}, \pointToAppendix{\ref{appendix:sim2real}}{A}, \pointToAppendix{\ref{appendix:mic-hrda}}{B}), including adaptation scenarios not tackled by prior works.
\noindent
In summary, our contributions are,
\begin{itemize}
    \vspace{-0.1in}
    \item We are the first to demonstrate that saliency guided image warping enhances backbone features and improves model adaptability for unsupervised domain adaptation (Section~\ref{subsec:dass-results} and \pointToAppendix{\ref{appendix:addl_analysis}}{E}).
    \vspace{-0.1in}
    \item Our approach is agnostic to the adaptation task (detection or segmentation), target domain (night, bad weather), saliency guidance, model architecture, and domain adaptation algorithm. 
    \vspace{-0.1in}
    \item Our approach is efficient, with train-time image warping and feature unwarping but no test-time warping, resulting in minimal training overhead and no additional inference latency (\pointToAppendix{\ref{appendix:mic-hrda}}{B}).
    \vspace{-0.1in}
    \item Our approach improves domain adaptation across various lighting (\textbf{\txtgreen{+3.7}} mAP50 for BDD100K Day $\rightarrow$ Night detection), weather conditions (\textbf{\txtgreen{+6.1}} mAP50 for BDD100K Clear $\rightarrow$ DENSE Foggy detection), and geographies (\textbf{\txtgreen{+6.3}} mIoU for Cityscapes $\rightarrow$ ACDC semantic segmentation).
\end{itemize}

\section{Related Work}
\label{sec:related-work}

\begin{figure*}[t!]
    \centering
    \includegraphics[width=0.9\textwidth]{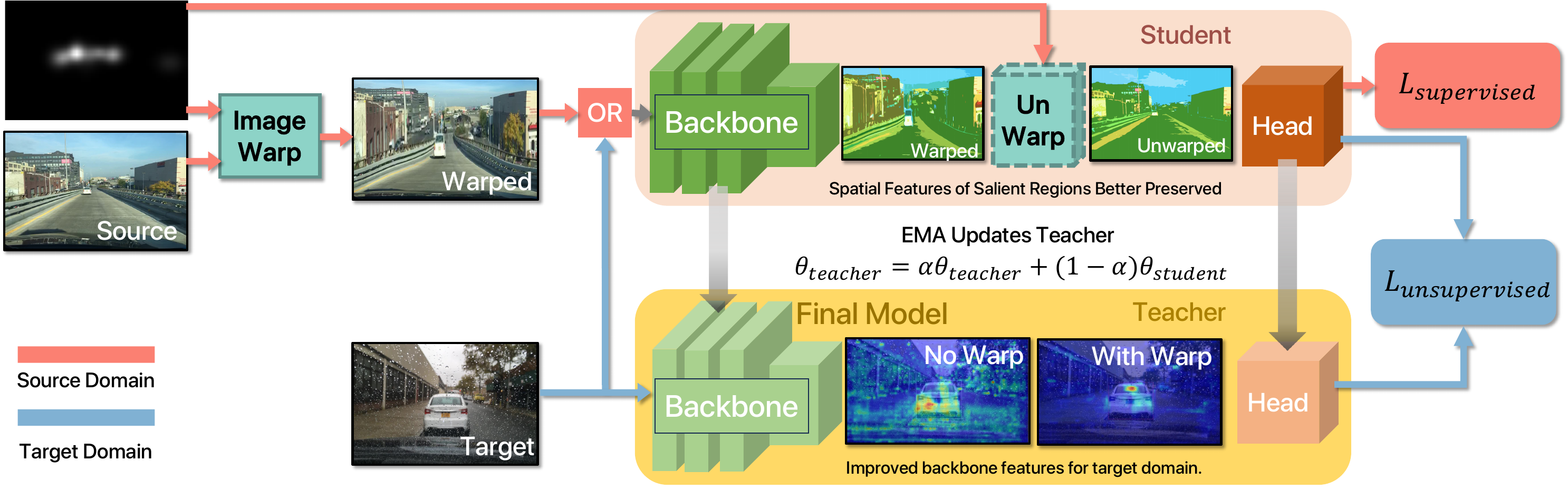}
    \vspace{-0.05in}
    \caption{\textbf{Saliency Guided Image Warping for Unsupervised Domain Adaptation}. Consider the standard UDA framework (See Sections~\ref{subsec:overview} and~\ref{subsec:da}) with supervised pre-training and unsupervised self-training phases (as shown here). The components of warping and unwarping are marked in \colorbox{ourlightcyan}{\textbf{\textcolor{black}{cyan}}}. We \textbf{warp} images using saliency guidance to oversample salient image regions, encouraging improved feature learning for the backhone. Our instance-level saliency guidance oversamples object regions, showing better performance compared to Static Prior~\cite{thavamani2021fovea, thavamani2023learning} and Geometric Prior~\cite{ghosh2023learned}. We \textbf{unwarp} features before predicting labels, ensuring that our labels are never warped, and the UDA losses of the employed algorithm (e.g., \cite{hoyer2022daformer, kennerley20232pcnet, hoyer2022hrda, hoyer2023mic, li2022cross, deng2021unbiased, zhang2021prototypical}) remain unmodified. We do not warp or unwarp at test time.
    }
    \label{fig:workflow}
    \vspace{-0.2in}
\end{figure*}

\noindent
\textbf{Saliency Guided Image Warping.} Non-uniform Spatial Transformations were traditionally used to correct image distortions arising from camera acquisition~\cite{wolberg1990digital, beier1992feature}. Recently, spatial transformation methods~\cite{jaderberg2015spatial,recasens2018learning} for learned models introduce intentional image distortions to enhance performance. Learning to Zoom~\cite{recasens2018learning} that over-sampling task-specific salient regions is useful, as visual recognition requires combining information from different spatial resolutions. Building on this paradigm, warping for other tasks like detection~\cite{thavamani2021fovea, thavamani2023learning} and segmentation~\cite{thavamani2023learning} has been explored. For e.g., Fovea~\cite{thavamani2021fovea} and Two Plane Prior~\cite{ghosh2023learned} designed `salient' priors to guide warping, introducing static \& temporal, and geometric saliency guidance, respectively.

Our approach stands outs in the following ways: \textit{(a)} Unlike many UDA methods that indiscriminately process object and background, we introduce an instance-level saliency guidance that targets regions containing objects. \textit{(b)} Unlike previous image warping efforts focused on supervised settings, our novel training strategy and saliency guidance tailor image warping for unsupervised domain adaptation. \textit{(c)} Unlike prior UDA works using augmentations such as resizing~\cite{hoyer2022hrda}, which are memory-intensive, or supervised warping works~\cite{thavamani2021fovea,ghosh2023learned,thavamani2023learning} that perform test-time warping and increase inference latency, our in-place warping incurs minimal memory and latency overhead during training and no additional latency during testing.


\noindent
\textbf{Image Transformations for Domain Adaptation.}  Transformations like flipping, cropping, blurring, color jittering, grayscaling, and contrast adjustment are common perturbations in domain adaptation~\cite{tsai2018learning, yang2018denseaspp, yang2020fda, mei2020instance, sakaridis2020map, zhang2021prototypical, tranheden2021dacs, deng2021unbiased, wang2021domain, wu2021dannet, li2022cross, hoyer2022daformer, hoyer2022hrda, he2022cross, hoyer2023mic, kennerley20232pcnet}. They increase source domain diversity and narrow the gap to target domain. For example, 2PCNet~\cite{kennerley20232pcnet} introduces a domain-specific Night Augmentation (NightAug) to make day images resemble night images. To the best of our knowledge, in-place image warping as a domain adaptive perturbation has not been explored previously. Our approach is agnostic to domain adaptation algorithms and can be seamlessly combined with other augmentation strategies.


\section{Method}

\subsection{Overview}
\label{subsec:overview}

\noindent
 We perform unsupervised domain adaptation (UDA), i.e. source domain $\mathcal{X}_{a}$ includes labeled data, i.e., $X_{a} = \{ \ (x_a, y_a) \in \mathcal{X}_{a} \ \}$ and target domain $\mathcal{X}_{b}$ includes unlabeled images, i.e., $X_{b} = \{ \ x_b \in \mathcal{X}_{b} \ \}$.

 The core of our approach lies in the operations of ``image warping'' and ``feature unwarping''. In a standard UDA framework~\cite{hoyer2022daformer, kennerley20232pcnet}, self-training involves a student and teacher network trained on different image augmentations or views~\cite{caron2021emerging, grill2020bootstrap} with an unsupervised loss following a supervised pre-training phase. In our method, the model \textit{views} ``warped'' images to encourage feature learning in oversampled regions. Although our model processes warped images, it predicts labels as if the image was never warped, accomplished through feature ``unwarping''~\cite{thavamani2023learning}. This ensures that ground truth or teacher labels remain unwarped, allowing unsupervised losses (e.g., from~\cite{hoyer2022daformer, kennerley20232pcnet, hoyer2022hrda, hoyer2023mic, li2022cross, deng2021unbiased, zhang2021prototypical}) to be used without modification. The ``warp'' and ``unwarp'' operations are employed during domain adaptation to improve backbone feature learning. The operations are not applied at test time, ensuing no added inference latency. Figure~\ref{fig:method-overview} illustrates the mechanism of Saliency Guided Warping for domain adaptation.

In Section~\ref{subsec:warping}, we discuss how to warp and unwarp images. Section~\ref{subsec:saliency-choice}, we discuss strategies of deciding which regions to sample to encourage improved feature learning. Section~\ref{subsec:da} discusses how to employ the pair of operations during domain adaptation.

\subsection{Image Warping and Feature Unwarping}
\label{subsec:warping}

\noindent
We employ the saliency guided sampling mechanism from~\cite{recasens2018learning}. However, other methods, such as thin plate spline transformations~\cite{jaderberg2015spatial, ehteshami2022salisa}, can also be used.

\noindent 
\textbf{Image Warping.} Following common practice in saliency guided image warping~\cite{beier1992feature, recasens2018learning, thavamani2021fovea} where warp $\mathcal{T}$ is defined through a backward map, we consider inverse transformation $\mathcal{T}^{-1}_{S}$ parameterized by saliency $S$ such that,

\vspace{-0.1in}
\begin{equation}
    I'(\mathbf{u}) = W_{\mathcal{T}}(I) = I(\mathcal{T}_{S}^{-1}(\mathbf{u}))
\end{equation}

\noindent
a resampled $I'$ for every location $\mathbf{u}$ is obtained by using warp $W_{\mathcal{T}}$ given input image $I$ and saliency guidance $S$, which is a 2D map of the output image size. The rationale and choice of $S$ is described in Section~\ref{subsec:saliency-choice}. Warping incurs an additional 1.4 ms latency per image on a 4090 GTX GPU and has no additional learned parameters.

\noindent
\textbf{Feature Unwarping.} If we operate in the space of warped images, our predictions would also be warped. However, we do not wish to warp the labels to match the warped input image. Instead, we perform feature unwarping on the spatial backbone features and train the model without warping labels or unwarping predictions like~\cite{thavamani2021fovea, ghosh2023learned}, and the losses remain unchanged. This makes our warping task-agnostic. However, $\mathcal{T}^{-1}$ has no closed form inverse~\cite{thavamani2021fovea, thavamani2023learning}. Therefore, we adopt the strategy from LZU~\cite{thavamani2023learning}, which approximates $\mathcal{T}^{-1}$ with $\widetilde{\mathcal{T}}^{-1}$, a piecewise tiling of \textit{invertible} bilinear maps, ensuring that both $\widetilde{\mathcal{T}}$ and $\widetilde{\mathcal{T}}^{-1}$ exist):

\vspace{-0.2in}
\begin{equation}
{\mathcal{T}}(\mathbf{u}) \approx 
\widetilde{\mathcal{T}}(\mathbf{u})= 
    \begin{cases}\widetilde{\mathcal{T}}_{i j}(\mathbf{u}) & \text { if } \mathbf{u} \in \operatorname{Range}\left(\widetilde{\mathcal{T}}_{i j}^{-1}\right) \\ 0 & \text { else }
    \end{cases}
\end{equation}

\noindent
We then unwarp features using the approximation $\widetilde{\mathcal{T}}$ (the closed-form inverse of $\widetilde{\mathcal{T}}^{-1}$) as our \textit{backward map}, since the roles of output and input coordinates are switched during unwarping. Unwarping incurs 4.2 ms latency per image on a 4090 GTX GPU and does not introduce any additional learned parameters. Figure~\ref{fig:warp-unwarp-error} shows that while warping and unwarping are lossy, the information loss is minimal.

\noindent
\textbf{Additional Considerations.} 
We restrict the space of allowable warps for $\mathcal{T}^{-1}$ to preserve axis alignment. In this approach, entire rows or columns are `stretched' or `compressed,' which are shown by~\cite{thavamani2021fovea, ghosh2023learned, thavamani2023learning} to perform better,

\begin{equation}
    \mathcal{T}_{S, u}^{-1}(u) = \frac{\int_{u'} S_{u}(u') k_{u}(u', u) u' du'}{\int_{u'} S_{u}(u') k_{u}(u, u') du'}
\end{equation}

\begin{equation}
    \mathcal{T}_{S, v}^{-1}(v) = \frac{\int_{v'} S_{v}(v') k_{v}(v', v) v' dv'}{\int_{v'} S_{v} (v') k_{v}(v, v') dv'}
\end{equation}

\noindent
where $k_{u}$ and $k_{v}$ are Gaussian kernels and $\mathbf{u} = (u, v)$ represent the two image axes in pixel space. $S_{u}$ and $S_{v}$ are the marginalized versions of $S$ along $u$ and $v$ axes, respectively.

\subsection{Designing Saliency Guidance}
\label{subsec:saliency-choice}

\begin{figure}[t!]
    \centering
    \includegraphics[width=\linewidth]{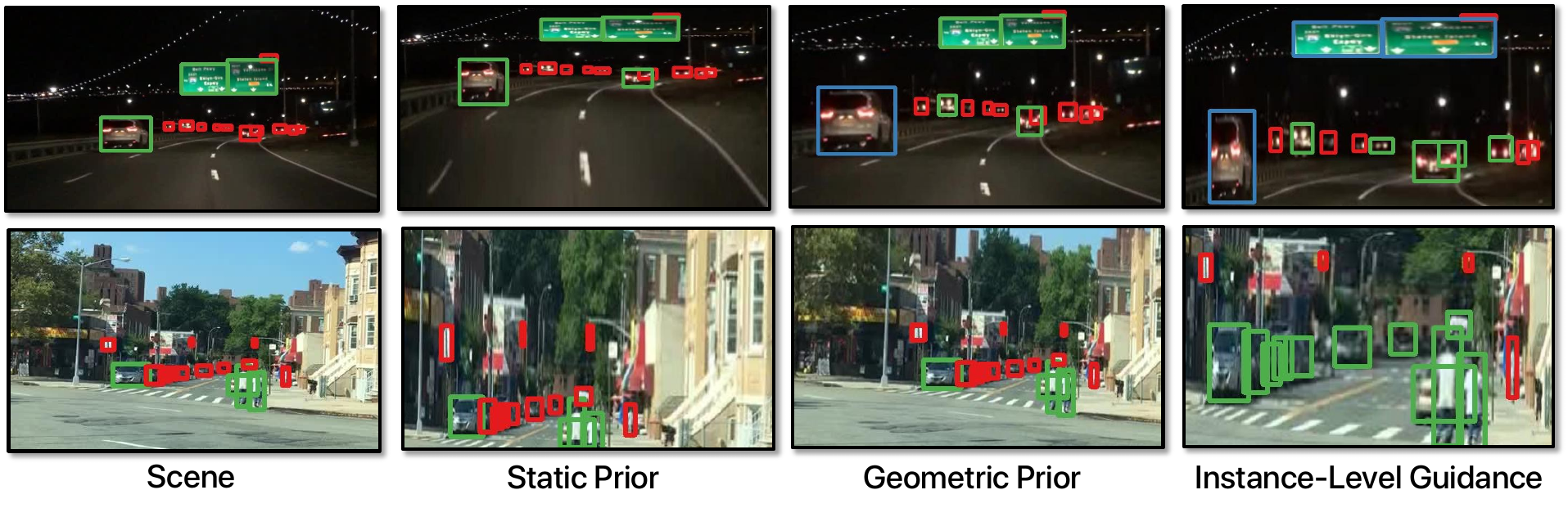}
    \vspace{-0.15in}
    \caption{ \textbf{Image Warping with Different Saliency Guidance Functions.} In-place warping follows a zero-sum pixel constraint: enlarging one region necessitates shrinking another. Bounding boxes mark \colorbox{ourmediumred}{\textcolor{white}{small}},  \colorbox{ourmediumgreen}{\textcolor{black}{medium}}, and \colorbox{ourmediumblue}{\textcolor{white}{large}} objects. Our instance-level saliency guidance oversamples objects and undersamples the background. In contrast, Static Prior fails when object locations do not align with the dataset's average object location, while Geometric Prior fails for small objects not near the vanishing point.}
    \vspace{-0.25in}
    \label{fig:teaser2}
\end{figure}

\noindent
Saliency ($S$) determines which regions of an image are sampled more during warping~\cite{recasens2018learning}. We explore three saliency functions: two based on approximate guidance from prior work~\cite{thavamani2021fovea, thavamani2023learning, ghosh2023learned}, and a novel instance-level saliency guidance that we propose. Since images are warped {\it{in-place}}, expanding some regions necessitates reducing others (Figure~\ref{fig:teaser2}). Only instance-level saliency guidance \textit{oversamples objects} and \textit{undersamples the background}.

\noindent
\textbf{Approximate Saliency Guidance.} The Geometric Prior~\cite{ghosh2023learned} saliency guidance is a function of the dominant vanishing point in the scene, oversampling objects close to the vanishing point, which are usually small due to their distance from the viewpoint. However, regions near the vanishing point may lack object, and not all objects may be sufficiently oversampled (Figure \ref{fig:teaser2}). 

The Static Prior~\cite{thavamani2021fovea, thavamani2023learning} assumes that object distribution remains consistent at test time, which is often untrue due to domain shift. The incorrect assumption results in a saliency map targeting average object locations, which may fail to accurately represent specific target images where object positions deviate from these averages (Figure \ref{fig:teaser2}). 

Moreover, warping during test time, after adaptation, not only increases inference latency but is also ineffective (See Section~\ref{subsec:ablation}). Essentially, approximate saliency guidance ignores groundtruth object labels and fails to focus on salient objects, which is crucial for improving adaptation.


\noindent
\textbf{Instance-level Saliency Guidance.} 
We propose Instance-level Saliency Guidance which leverages ground truth labels in the source domain (Figure \ref{fig:teaser2}). During training, source image $x_a$ has corresponding labels $y_a$ available. For simplicity, we assume the labels are bounding boxes.

Given an image, we aim to \textit{oversample object pixels}, such as cars, and \textit{undersample background pixels}, such as roads. Following~\cite{bi2018accurate, thavamani2021fovea}, we use kernel density estimation to model the bounding box saliency as a sum of Gaussians,

\begin{figure}[t!]
    \centering
    \includegraphics[width=\linewidth]{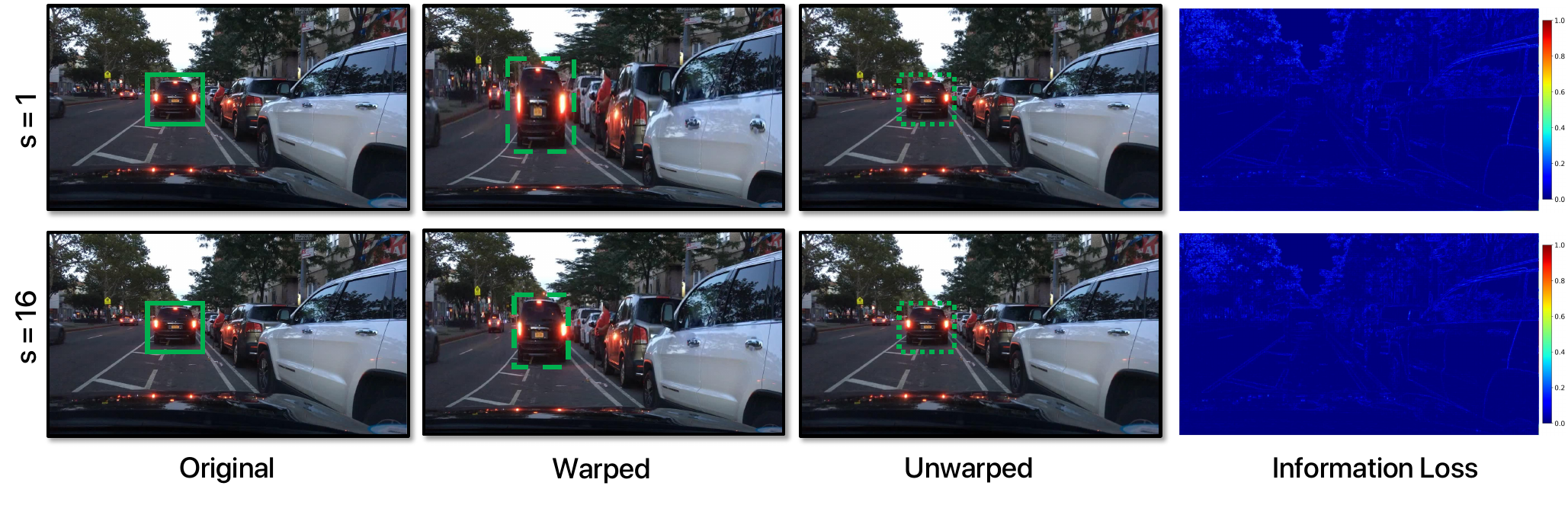}
    \vspace{-0.15in}
    \caption{\textbf{Warping and Unwarping.} Shown is a warped image with two different saliency scales $s = \{1, 16\}$. A higher saliency scale implies less intense warping. While unwarping is applied to features, here we apply it to the warped image for illusration. Although saliency guided warping and unwarping are lossy, we observe that the error between the original image and warped-then-unwarped image is very low, indicating minimal information loss.}
    \vspace{-0.2in}
    \label{fig:warp-unwarp-error}
\end{figure}



\vspace{-0.1in}
\begin{equation}
S=\sum_{\left(c_i, w_i, h_i\right) \in y_a} \mathcal{N}\left(c_i, s\left[\begin{array}{cc}
w_i & 0 \\
0 & h_i
\end{array}\right]\right)
\end{equation}

\noindent
where $c_i$, $w_i$ and $h_i$ represent the center, width, and height of the bounding box, $\mathcal{N}(\mu, \Sigma)$ is the normal distribution and $s$ is the saliency scale. Figure~\ref{fig:warp-unwarp-error} shows the effect of different saliency scales, where a higher scale results in more warping areas but less warping intensity. The expansion factor $f$ ($s = \frac{P}{f}$) reflects warping intensity, where $P$ is a constant.


Intuitively, for dataset with many small objects, high warping intensity is helpful. However, if the dataset contains mostly large objects, warping would distort them and harm performance. In this case, the warping intensity should be reduced. We define $f$ following this intuition as:

\vspace{-0.1in}
\begin{equation}
f = 2^{\max\left(\left\lfloor \displaystyle\log_2\left( 
\sum_{\psi \in \Psi} \frac{\psi_{i}}{\psi_{i+1}}
\right) \right\rfloor, 0\right)}
\end{equation}

\noindent
where $\Psi$ is the fraction of object size corresponding to an interval between size thresholds $t_{i}$ and $t_{i+1}$. In our case, we consider $\Psi = \{ \psi_{small}, \psi_{medium}$, $\psi_{large} \}$ to represent the percentage of small, medium, and large objects in the dataset, as per COCO~\cite{lin2014microsoft} object size thresholds. In the rare case of having only large objects in the dataset, $f$ approaches 1 (minimal warping). Conversely, with only small objects, $f$ tends to a large value (very intense warping).

We clip $S$ between a lower bound of 0, to prevent negative saliency values that lead to holes, and an upper bound $U$, to discourage extreme warping. Figure~\ref{fig:warp-unwarp-error} shows that a higher saliency scale  $s$ results in less intense warping.

When segmentation masks are available but bounding boxes are not, we convert the masks into `from-seg' boxes. This does not significantly impact performance (See Section~\ref{subsec:ablation} and \pointToAppendix{\ref{appendix:addl_tech}}{G}). 

\begin{figure*}[t!]
    \centering
    \includegraphics[width=0.8\textwidth]{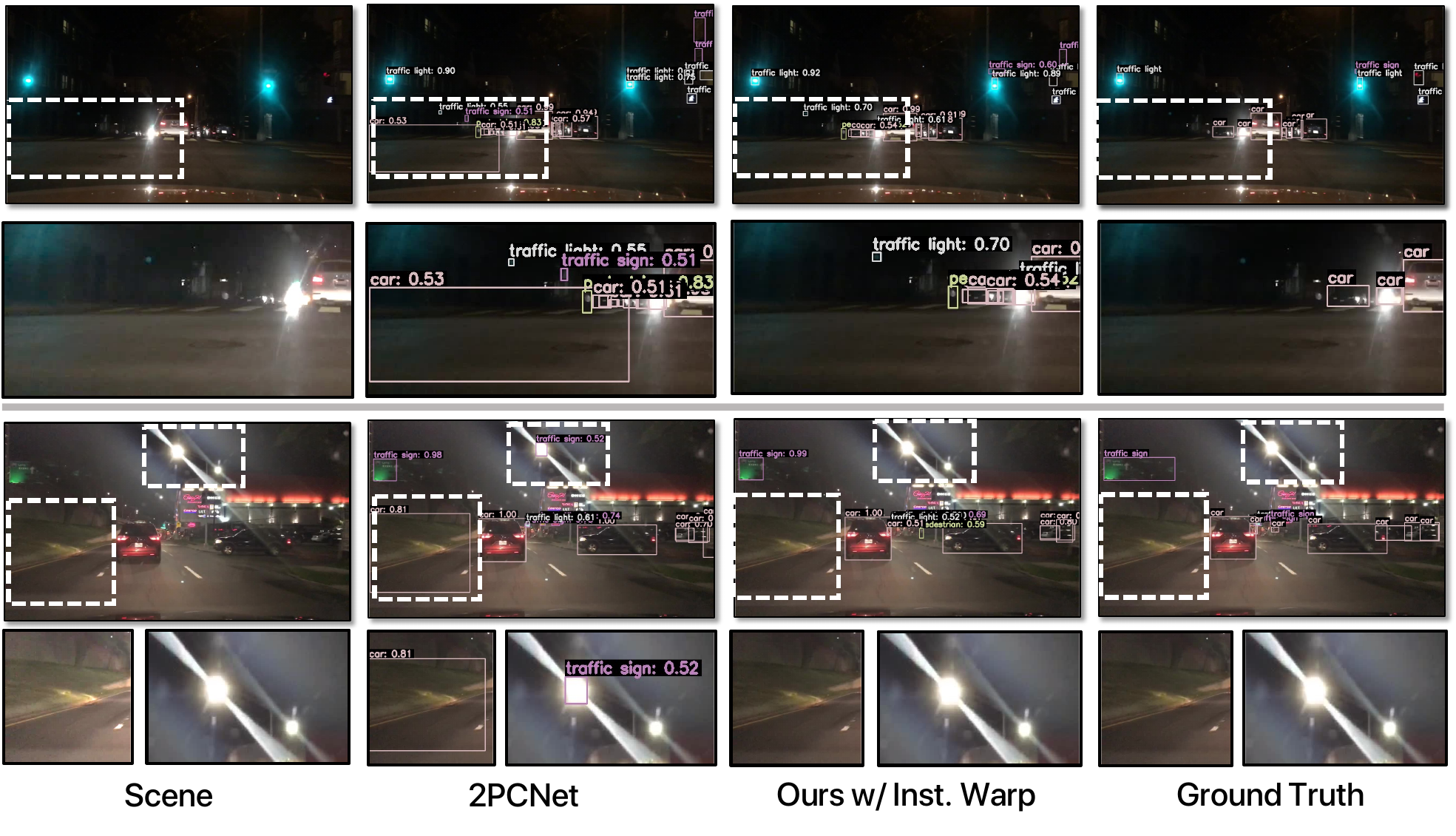}
    \vspace{-0.1in}
    \caption{\textbf{Domain Adaptive Object Detection (BDD100K Day $\rightarrow$ Night).} Our warping improves scene understanding by enabling 2PCNet to better distinguish between different objects and between objects and backgrounds. In contrast, 2PCNet~\cite{kennerley20232pcnet} baseline misclassifies roads as cars and streetlights as traffic signs.}
    \label{fig:det_day_night}
    \vspace{-0.05in}
\end{figure*}

\begin{figure*}[t!]
    \centering
    \includegraphics[width=0.8\textwidth]{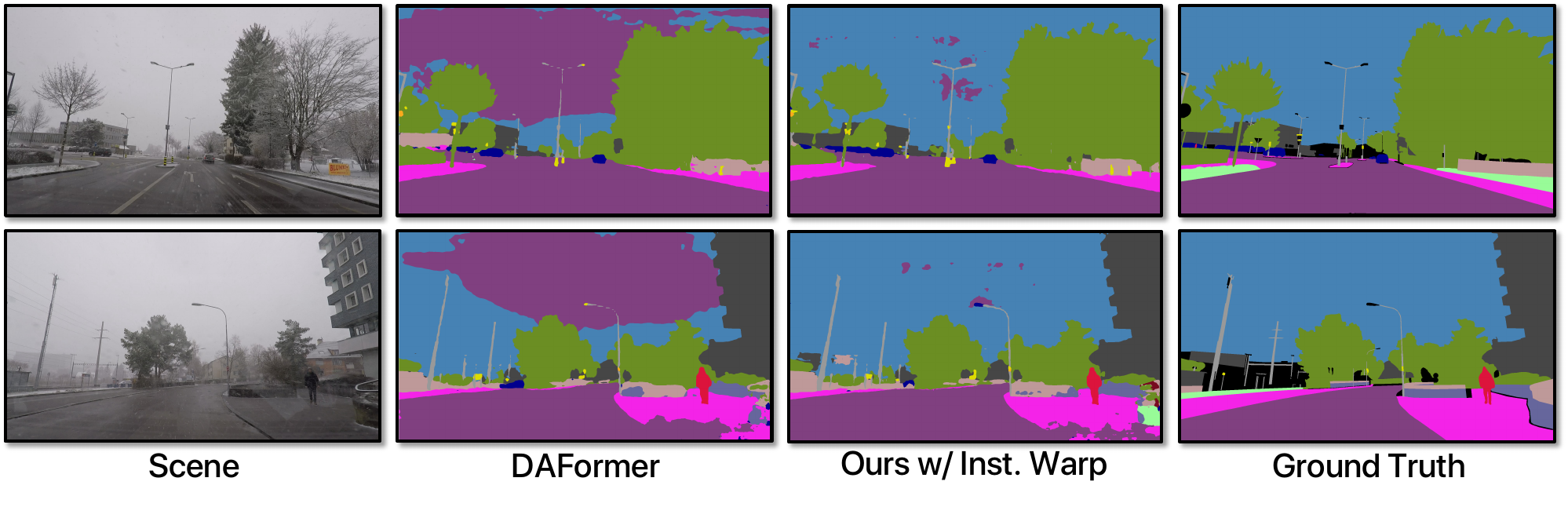}
    \vspace{-0.1in}
       \caption{\textbf{Domain Adaptive Semantic Segmentation (Cityscapes $\rightarrow$ ACDC).}  Our warping improved recognition of backgrounds (e,g., sky, sidewalks, and buildings), and foreground objects (e.g., traffic lights, traffic signs), indicating enhanced scene understanding.}
    \label{fig:seg_acdc}
    \vspace{-0.15in}
\end{figure*}

\subsection{Warping for Unsupervised Domain Adaptation}
\label{subsec:da}

\noindent
We first discuss the standard domain adaptation framework, followed by our modifications to incorporate warping and unwarping into this framework.

\noindent
\textbf{Domain Adaptation Framework.} The standard framework follows a student-teacher self-training approach. It begins with a supervised pre-training stage where a model $\hat{y}_a = F_{\theta}(x_a)$ is trained using supervised loss on the source domain, 
$\mathcal{L}_{sup}(y_a, F_{\theta}(x_a)).$    
Next, in the unsupervised adaptation stage, the model is adapted to the target domain $\mathcal{X}_b$ using dataset $X_{b}$ through an additional unsupervised loss. Generally, a student $F_{s, \phi}$ and teacher $F_{t, \kappa}$ (initialized using $F_{\theta}$) are trained jointly through self-training with psuedo-labels~\cite{kennerley20232pcnet, hoyer2022daformer}. The main insight of self-training is that the student and teacher branches are trained with different image augmentations or views, and their predicted labels should match each other. Following \cite{hoyer2022daformer,hoyer2022hrda,hoyer2023mic,kennerley20232pcnet}, we update the teacher parameters, as DINO~\cite{caron2021emerging} and Mean Teachers~\cite{tarvainen2017mean} have shown that this approach leads to better performance compared to traditional distillation. While training the student, a supervised loss is used to maintain performance on the source domain $\mathcal{X}_a$ in a supervised branch -- thus loss is
$\mathcal{L}_{sup}(y_a, F_{s, \phi}(x_a)) + \lambda\mathcal{L}_{unsup}(F_{t, \kappa}(x_b), F_{s, \phi}(x_b))$.


\noindent Meanwhile, the teacher is updated as the exponential moving average (EMA) of the student like\cite{hoyer2022daformer,hoyer2022hrda,hoyer2023mic,kennerley20232pcnet}.

\begin{table}[t]
\centering
\caption{\textbf{Adverse Weather: BDD100K Clear $\rightarrow$ DENSE Foggy Object Detection.} Our method achieves notable improvements in \textbf{\txtgreen{+6.1}} mAP, \textbf{\txtgreen{+6.1}} mAP50, \textbf{\txtgreen{+9.6}} mAP75 and \textbf{\txtgreen{+9.7}} mAPl over 2PCNet. Instance-Level Guidance is superior to other functions.}
\resizebox{\columnwidth}{!}{%
\begin{tabular}{l|cccccc}
\toprule
Method & mAP & \textbf{mAP50} & mAP75 & mAPs & mAPm & mAPl \\ \hline
2PCNet          & 27.0        & 45.0          & 30.1          & 0.0          & 15.4        & 35.5       \\ \hline
+ Ours w/ Sta. Prior      &  28.6         &  48.5            &   31.7       &   0.0       & \textbf{19.6}      &  34.0   	\\
+ Ours w/ Geo. Prior      &   27.1       &   45.8           &     29.5     &   0.0       &  15.2     &   37.5        \\
+ Ours w/ Inst. Warp         & \textbf{33.1}     & \textbf{51.1}    & \textbf{39.7}          & 0.0          & 14.5         & \textbf{45.2}   \\ \bottomrule
\end{tabular}%
}
\vspace{-0.1in}
\label{tab:bdd_dense_foggy}
\end{table}

\begin{table}[t]
\centering
\caption{\textbf{Adverse Weather: BDD100K (Clear $\rightarrow$ Rainy) Object Detection. }  Tested on BDD100k Rainy Val. Instance-Level Guidance improves 2PCNet~\cite{kennerley20232pcnet} by \textbf{\txtgreen{+3.0}} mAP50. }
\resizebox{\columnwidth}{!}{%
\begin{tabular}{l|cccccc}
\toprule
Method & mAP & \textbf{mAP50}  & mAP75  &	mAPs  &	mAPm  &	mAPl    \\ \hline
2PCNet         & 25.8   & 48.4  & 24.4 &	10.9 &	34.1 &	49.0                                    \\   \hline
+ Ours w/ Sta. Prior     &   26.8     & 50.6  & \textbf{24.9}	 & 11.4	 & 35.9	  & 49.6                \\ 
+ Ours w/ Geo. Prior        &  26.9     & 49.9   & 24.5 &  \textbf{11.8} &  35.6 & \textbf{50.1}        \\ 
+ Ours w/ Inst. Warp      & \textbf{27.1}  & \textbf{51.4}  & 24.6  & 11.0  & \textbf{36.7}  & 49.0              \\ 
\bottomrule

\end{tabular}%
}
\vspace{-0.1in}
\label{tab:bdd100k_clear2rainy}
\end{table}

\noindent
\textbf{Warping and Unwarping for Domain Adaptation.} 
Our warping mechanism $W$ operates \textit{only on source domain images while training}, i.e. it is employed in the supervised pre-training stage and in the student model during the unsupervised stage when source images are used. Consider the recognition model $F$ as a combination of backbone $B$ and prediction heads $H$, i.e. $F(x) = H(B(x))$. We warp the image and then unwarp the feature maps from the backbone $B$ to compose the final output,

\vspace{-0.15in}
\begin{equation}
    F'(x_a) = H(W_{\widetilde{\mathcal{T}}^{-1}}(B(W_{\mathcal{T}}(x_{a}))))    
\end{equation}

\noindent
following~\cite{thavamani2023learning} for a source image $x_{a}$. Thus, losses are modified for the supervised pre-training stage as $\mathcal{L}_{sup}(y_a, F'_{\theta}(x_a))$ and for the unsupervised training stage as, 

\vspace{-0.2in}
\begin{equation}
    \mathcal{L}_{sup}(y_a, F'_{s, \phi}(x_a)) + \lambda\mathcal{L}_{unsup}(F_{t, \kappa}(x_b), F_{s, \phi}(x_b))
\end{equation}

 
Instance-level saliency requires ground truth labels, making it unfeasible for warping target images during UDA. However, we experimented by warping target images (during training and testing) using Static~\cite{thavamani2021fovea,thavamani2023learning} and Geometric~\cite{ghosh2023learned} saliency guidance, which do not require ground truth. However, this degrades performance (See Table~\ref{table:src-tgt-ablation}). We hypothesize that warping both source and target images hinders the adaptability of the student model. Therefore, We warp only source domain images, while target domain images remain warped. This encourages the student model to learn to predict in both the warped and unwarped image spaces, as it is exposed to both types of views. In contrast, the teacher model only observes the unwarped views. The disparity between the teacher's and the student's view can be regarded as a form of data augmentation. We leave the analysis of the effect of this teacher-student disparity, which has also been explored in works like~\cite{caron2021emerging, grill2020bootstrap, hoyer2022daformer, hoyer2023mic, hoyer2022hrda, kennerley20232pcnet}, for future work. Finally, we do not warp during test time because it introduces additional latency and our ablation study (Table~\ref{table:src-tgt-ablation}) shows that it does not improve performance.



\begin{table}[t]
\centering
\caption{\textbf{Bad Lighting: BDD100K (Day $\rightarrow$ Night) Object Detection.} Tested on BDD100k Night Val. Instance-Level Guidance improves 2PCNet~\cite{kennerley20232pcnet} by \textbf{\txtgreen{+3.7}} mAP50. } 
\resizebox{\columnwidth}{!}{%
\begin{tabular}{l|cccccc}
\toprule
Method  & mAP  & \textbf{mAP50}  &  mAP75	&  mAPs	& mAPm	& mAPl  \\ \hline
2PCNet       & 23.5   & 46.4   & 21.1 & 9.5 &  26.0 &  43.3      \\ \hline 
+ Ours w/ Sta. Prior        & 25.3 & 49.4     & 22.4 & \textbf{10.3} & 27.4 & 45.1       \\ 
+ Ours w/ Geo. Prior          & \textbf{25.6}   & 49.7   & \textbf{22.9} & 10.0 & \textbf{27.7}  & 45.7      \\ 
+ Ours w/ Inst. Warp   & \textbf{25.6}   & \textbf{50.1}     & 22.6  & 9.8 & 27.6 & \textbf{46.6}    \\ 
\bottomrule
\end{tabular}%
}
\vspace{-0.1in}
\label{tab:bdd100k_day2night}
\end{table}

\begin{table}[t]
\centering
\caption{\textbf{Changing Geography: BDD100K Clear $\rightarrow$ ACDC Object Detection.} Tested on ACDC Val. Instance-Level Guidance improves 2PCNet~\cite{kennerley20232pcnet} by \textbf{\txtgreen{+1.3}} mAP50. }
\resizebox{\columnwidth}{!}{%
\begin{tabular}{l|cccccc}
\toprule
Method  & mAP	& \textbf{mAP50}	& mAP75	 & mAPs	& mAPm	& mAPl \\ \hline
2PCNet          & 16.6 & 30.8   & 14.9  & 6.0  & 24.8  & 26.6    \\   \hline
+ Ours w/ Sta. Prior        & 17.3   & 31.0   & 16.7  & 6.2  & 25.9  &  29.5       \\ 
+ Ours w/ Geo. Prior         &  16.6   & 30.3     &  15.4  & 5.2   &  25.1  & 28.7     \\ 
+ Ours w/ Inst. Warp    &  \textbf{17.9}   & \textbf{32.1}    & \textbf{18.2}   & \textbf{6.8}  &  \textbf{26.7}   &  \textbf{32.1}      \\ 
\bottomrule
\end{tabular}%
}
\vspace{-0.1in}
\label{tab:bdd100k_acdc}
\end{table}

\noindent
\textbf{Base Algorithm and Losses.} We use DAFormer~\cite{hoyer2022daformer} and 2PCNet~\cite{kennerley20232pcnet} as our base segmentation and detection adaption algorithms, respectively. We directly adopt their losses for $L_{sup}$ and $L_{unsup}$ without modification. Our method is task-agnostic (segmentation or detection). Therefore, other algorithms~\cite{hoyer2022hrda, hoyer2023mic, li2022cross, deng2021unbiased, zhang2021prototypical, wang2023balancing} are equally applicable.


\noindent
\textbf{Test Time Inference.} We employ the teacher model $F_{t}$ at test time. Our method does not require warping or unwarping during testing. This contrast with image warping techniques proposed in the supervised setting~\cite{thavamani2021fovea, ghosh2023learned, thavamani2023learning}, which incur overheads from test-time warping and require additional information (like vanishing points~\cite{ghosh2023learned}) to generate saliency maps for warping at test time.

\noindent
\textbf{Supervised Setting.} Our training method can also be applied in supervised learning scenario by following the supervised pre-training steps to train a model $F'$. That is, omitting the unsupervised training stage completely and only performing supervised pre-training stage -- then removing the warping and unwarping layers completely from the trained model. This is unlike prior work~\cite{recasens2018learning, thavamani2021fovea, ghosh2023learned, thavamani2023learning} which also advocates warping at test time. With our warping on the source domain, model performance improves on the target domain $\mathcal{X}_{b}$ even without adaptation (See Table~\ref{tab:cs2acdc} and \ref{tab:cs2dz}). For Source Domain Results, see \pointToAppendix{\ref{appendix:supervised}}{C}.



\begin{table*}[!t]
\centering
\caption{\textbf{Cityscapes $\rightarrow$ ACDC Semantic Segmentation.} Tested on ACDC Val and Test-splits. Our method is superior, with or without Adaptation, indicating better generalization of backbone. We improves DAFormer by \textbf{\txtgreen{+6.3}} mIoU on Test-split.}
\resizebox{\textwidth}{!}{%
\begin{tabular}{l|c|ccccccccccccccccccccc}
\toprule
Method         & \textbf{mIoU} & road & \begin{tabular}[c]{@{}c@{}}side\\ walk\end{tabular} & building & wall & fence & pole & \begin{tabular}[c]{@{}c@{}}traffic\\ light\end{tabular} & \begin{tabular}[c]{@{}c@{}}traffic\\ sign\end{tabular} & \begin{tabular}[c]{@{}c@{}}vege-\\ tation\end{tabular} & terrain & sky  & person & rider & car  & truck & bus  & train & \begin{tabular}[c]{@{}c@{}}motor\\ cycle\end{tabular} & bike \\ \hline

\multicolumn{21}{c}{\textbf{ACDC Val Split: After Supervised Pre-Training}}                                                                                                                                                                                                                                                                                                                                                                                                                                    \\ \hline
DAFormer~\cite{hoyer2022daformer}                           & 55.8                  & 77.6          & \textbf{57.4}                                       & 77.9          & 39.4          & 30.4          & 48.3          & 54.5                                                    & \textbf{46.4}                                          & 70.0                                                   & \textbf{37.9} & 72.2          & 52.6          & 32.0          & 81.4          & 72.7          & 73.7          & 70.4          & 37.6                                                  & 27.3          \\  \hline
+ Ours w/ Sta.                                     & 55.8                   & 77.5          & 55.8                                                & 76.7          & \textbf{39.8} & 28.7          & \textbf{48.6}          & 56.0                                                    & 43.9                                                   & 69.0                                                   & 37.5          & 73.0          & \textbf{53.6} & \textbf{32.9} & 81.1          & 71.8          & 72.6          & \textbf{82.5} & 36.0                                                  & 22.8          \\
+ Ours w/ Geo.                              & 55.7                 & \textbf{80.4} & 57.3                                                & \textbf{78.3} & 37.7          & 28.4          & 48.2          & 54.4                                                    & 44.8                                                   & \textbf{70.3}                                          & 37.4          & 75.1          & 52.8          & 30.5          & 80.7          & 75.6          & 72.8          & 65.8          & 38.9                                                  & 29.8          \\
+ Ours w/ Inst.                              & \textbf{56.8}  & 80.2          & 48.3                                                & 76.0          & 35.0          & \textbf{31.6} & \textbf{48.6} & \textbf{58.2}                                           & 45.3                                                   & 69.4                                                   & 36.9          & \textbf{77.2} & 53.2          & 26.2          & \textbf{82.0} & \textbf{77.6} & \textbf{80.0} & 69.4          & \textbf{39.7}                                         & \textbf{43.7} \\ \hline
\multicolumn{21}{c}{\textbf{ACDC Val Split: After Unsupervised Domain Adaptation}}                                                                                                                                                                                                                                                                                                                                                                                                                                                                                                                                                                                                                 \\ \hline
DAFormer~\cite{hoyer2022daformer}                              & 57.6                & 72.7          & 57.5                                                & \textbf{80.1} & 42.5          & 38.0          & 50.9          & 45.1                                                    & 50.0                                                   & 71.1                                                   & 38.5          & 67.0          & 56.0          & 29.9          & 81.8          & 76.6          & 78.9          & 79.9          & 40.8                                                  & 36.7          \\  \hline
+ Ours w/ Sta.                      & 59.1               & 73.8          & 56.8                                                & 79.1          & \textbf{45.2} & 39.6          & 50.0          & 48.3                                                    & 49.4                                                   & 71.0                                                   & \textbf{39.3} & 68.6          & 55.8          & \textbf{36.9} & 82.1          & 73.2          & 81.3          & \textbf{88.7} & 37.1                                                  & \textbf{47.4} \\
+ Ours w/ Geo.                         & 58.4                & 76.4          & \textbf{57.6}                                       & 79.9          & 42.2          & 39.0          & 51.7          & 46.4                                                    & 49.5                                                   & 72.0                                                   & 39.0          & 71.1          & 56.5          & 30.4          & 82.7          & 79.3          & 79.1          & 69.8          & 41.5                                                  & 44.7          \\
+ Ours w/ Inst.                        & \textbf{61.8}  & \textbf{82.9} & 56.1                                                & 79.8          & 44.6          & \textbf{40.3} & \textbf{52.7} & \textbf{60.8}                                           & \textbf{52.5}                                          & \textbf{72.0}                                          & 38.4          & \textbf{78.0} & \textbf{56.6} & 30.5          & \textbf{84.9} & \textbf{80.2} & \textbf{86.9} & 86.4          & \textbf{44.5}                                         & 45.8          \\ \hline
\multicolumn{21}{c}{\textbf{ACDC Test Split Results}} \\ \hline
ADVENT~\cite{vu2019advent}          & 32.7 & 72.9 & 14.3                                                & 40.5     & 16.6 & 21.2  & 9.3  & 17.4                                                    & 21.2                                                   & 63.8                                                   & 23.8    & 18.3 & 32.6   & 19.5  & 69.5 & 36.2  & 34.5 & 46.2  & 26.9                                                  & 36.1 \\
MGCDA~\cite{sakaridis2019guided}           & 48.7 & 73.4 & 28.7                                                & 69.9     & 19.3 & 26.3  & 36.8 & \textbf{53.0}                                                    & 53.3                                                   & \textbf{75.4}                                                   & 32.0    & 84.6 & 51.0   & 26.1  & 77.6 & 43.2  & 45.9 & 53.9  & 32.7                                                  & 41.5 \\
DANNet~\cite{wu2021dannet}          & 50.0 & \textbf{84.3} & \textbf{54.2}                                                & 77.6     & 38.0 & 30.0  & 18.9 & 41.6                                                    & 35.2                                                   & 71.3                                                   & 39.4    & \textbf{86.6} & 48.7   & 29.2  & 76.2 & 41.6  & 43.0 & 58.6  & 32.6                                                  & 43.9 \\ \hline
DAFormer~\cite{hoyer2022daformer}        & 55.4 & 58.4 & 51.3                                                & 84.0     & 42.7 & 35.1  & \textbf{50.7} & 30.0                                                    & 57.0                                                   & 74.8                                                   & 52.8    & 51.3 & 58.3   & \textbf{32.6}  & 82.7 & 58.3  & 54.9 & 82.4  & 44.1                                                  & 50.7 \\
+ Ours w/ Inst. & \textbf{61.7} & 83.0 & 53.2                                                & \textbf{85.5}     & \textbf{47.4} & \textbf{38.3}  & 46.0 & 51.4                                                    & \textbf{57.8}                                                   & 73.9                                                   & \textbf{56.2}    & 82.1 & \textbf{61.3}   & 32.3  & \textbf{85.5} & \textbf{69.0}  & \textbf{68.9} & \textbf{82.5}  & \textbf{46.7}                                                  & \textbf{52.0} \\ \toprule
\end{tabular}%
}
\vspace{-0.1in}
\label{tab:cs2acdc}
\end{table*}

\begin{table*}[t]
\centering
\caption{\textbf{Cityscapes $\rightarrow$ DarkZurich Semantic Segmentation}. Tested on DarkZurich Val and Test-splits. Our method is superior, with or without Adaptation, indicating better generalization of the backbone. We improves DAFormer by \textbf{\txtgreen{+3.3}} mIoU on Val-split. Our method is less effective (+0.6 mIoU) on the Test-split. Analyzing this discrepancy is difficult due to private test labels. However, we notice that test images have better lighting and less motion blur than val images. Possibly, the domain gap is smaller, and the baseline is already strong.}
\resizebox{\textwidth}{!}{%
\begin{tabular}{l|c|ccccccccccccccccccccc}
\toprule
Method         & \textbf{mIoU} & road & \begin{tabular}[c]{@{}c@{}}side\\ walk\end{tabular} & building & wall & fence & pole & \begin{tabular}[c]{@{}c@{}}traffic\\ light\end{tabular} & \begin{tabular}[c]{@{}c@{}}traffic\\ sign\end{tabular} & \begin{tabular}[c]{@{}c@{}}vege-\\ tation\end{tabular} & terrain & sky  & person & rider & car  & truck & bus  & train & \begin{tabular}[c]{@{}c@{}}motor\\ cycle\end{tabular} & bike \\ \hline

\multicolumn{21}{c}{\textbf{DarkZurich Val Split: After Supervised Pre-Training}}                                                                                                                                                                                                                                                                                                                                                                                                                                                                                                                                                                                      \\ \hline
DAFormer                                & 30.0                   & 90.7          & 52.7                                                & 45.5          & 18.4          & 39.0          & 38.4          & 27.5                                                    & 18.3                                                   & \textbf{59.0}                                          & 26.5          & 11.1          & 20.4          & 21.1          & 58.6          & -   & - & -   & 22.1                                                  & 20.0          \\  \hline
+ Ours w/ Sta.                            & 32.7                  & 91.8          & 57.2                                                & 47.3          & 19.3          & \textbf{46.2} & \textbf{40.1} & 21.0                                                    & 18.2                                                   & 53.5                                                   & \textbf{31.9} & 6.4           & 20.6          & 18.6          & \textbf{64.1} & -   & - & -   & 25.0                                                  & 26.4          \\
+ Ours w/ Geo.                            & 30.2                  & 89.9          & 50.5                                                & 45.9          & 20.5          & 36.8          & 39.1          & 21.0                                                    & \textbf{19.8}                                          & 54.7                                                   & 28.4          & 7.9           & 19.0          & \textbf{31.1} & 58.4          & -   & - & -   & 18.1                                                  & 33.4          \\
+ Ours w/ Inst.                          & \textbf{34.0}  & \textbf{91.9} & \textbf{59.4}                                       & \textbf{49.3} & \textbf{27.7} & 39.2          & 39.4          & \textbf{27.6}                                           & 15.8                                                   & 55.1                                                   & 29.9          & \textbf{11.4} & \textbf{20.7} & 9.2           & 61.9          & -   & - & -   & \textbf{27.9}                                         & \textbf{44.7} \\ \hline
\multicolumn{21}{c}{\textbf{DarkZurich Val Split: After Unsupervised Domain Adaptation}}                                                                      \\ \hline
DAFormer                              & 33.8                 & 92.0          & 66.7                                                & 47.3          & 25.9          & 50.8          & 38.0          & 24.6                                                    & 19.4                                                   & \textbf{62.2}                                          & 31.9          & 17.9          & 20.4          & \textbf{28.4} & 61.7          & -   & - & -   & 21.0                                                  & 34.3          \\  \hline
+ Ours w/ Sta.                             & 33.8             & \textbf{93.3} & 68.5                                                & 50.0          & 28.4          & \textbf{54.3} & \textbf{41.7} & 19.9                                                    & 22.7                                                   & 54.0                                                   & \textbf{41.5} & 14.1          & 22.9          & 7.2           & 62.6          & -   & - & -   & \textbf{24.6}                                         & \textbf{36.8} \\
+ Ours w/ Geo.                             & 33.3                 & 84.3          & 60.3                                                & \textbf{60.2} & 23.9          & 49.4          & 38.7          & 12.7                                                    & \textbf{25.0}                                          & 52.0                                                   & 39.5          & 20.8          & \textbf{37.1} & 16.2          & 64.6          & -   & - & -   & 21.1                                                  & 27.8          \\
+ Ours w/ Inst.                         & \textbf{37.1}  & 88.7          & \textbf{70.9}                                       & 60.1          & \textbf{42.1} & 49.6          & 39.8          & \textbf{47.9}                                           & 21.5                                                   & 49.8                                                   & 35.7          & \textbf{25.4} & 23.8          & 25.9          & \textbf{66.8} & -   & - & -   & 24.3                                                  & 32.4          \\  \hline

\multicolumn{21}{c}{\textbf{DarkZurich Test Split Results}} \\ \hline
ADVENT~\cite{vu2019advent} & 29.7 & 85.8 & 37.9 & 55.5 & 27.7 & 14.5 & 23.1 & 14.0 & 21.1 & 32.1 & 8.7 & 2.0 & 39.9 & 16.6 & 64.0 & 13.8 & 0.0 & 58.8 & 28.5 & 20.7 \\
MGCDA~\cite{sakaridis2019guided} & 42.5 & 80.3 & 49.3 & 66.2 & 7.8 & 11.0 & 41.4 & 38.9 & 39.0 & 64.1 & 18.0 & 55.8 & 52.1 & \textbf{53.5} & 74.7 & \textbf{66.0} & 0.0 & 37.5 & 29.1 & 22.7 \\
DANNet~\cite{wu2021dannet} & 44.3 & 90.0 & 54.0 & \textbf{74.8} & 41.0 & 21.1 & 25.0 & 26.8 & 30.2 & \textbf{72.0} & 26.2 & \textbf{84.0} & 47.0 & 33.9 & 68.2 & 19.0 & 0.3 & 66.4 & 38.3 & 23.6 \\ \hline
DAFormer~\cite{hoyer2022daformer} & 53.8 & 93.5 & 65.5 & 73.3 & 39.4 & 19.2 & \textbf{53.3} & 44.1 & 44.0 & 59.5 & \textbf{34.5} & 66.6 & 53.4 & 52.7 & \textbf{82.1} & 52.7 & 9.5 & 89.3 & 50.5 & 38.5 \\ 
+ Ours w/ Inst. & \textbf{54.4} & \textbf{94.2} & \textbf{68.1} & 72.6 & \textbf{43.2} & \textbf{23.6} & 51.0 & \textbf{48.1} & \textbf{48.5} & 60.6 & 33.6 & 54.8 & \textbf{55.5} & 49.0 & 79.8 & 51.1 & \textbf{13.2} & \textbf{90.2} & \textbf{53.2} & \textbf{43.4} \\ \toprule
\end{tabular}%
}  
\vspace{-0.15in}
\label{tab:cs2dz}
\end{table*}




\vspace{-0.05in}
\section{Experiments}

\noindent
We implement saliency-guided warping alongside a base domain adaptation algorithm, ensuring fair comparisons by using the same model, datasets, training schedules, hyperparameters, and seeds. We compare with 2PCNet~\cite{kennerley20232pcnet} for detection, and DAFormer~\cite{hoyer2022daformer}, HRDA~\cite{hoyer2022hrda} and MIC~\cite{hoyer2023mic} for segmentation. More details are provided in \pointToAppendix{\ref{appendix:addl_details}}{F}. 



\begin{table*}[t]
\caption{\textbf{Ablations}: (a) Studying the impact of hyperparamaters $P$ and $U$, same parameters used for all other experiments. (b) Studying the impact of balancing the object distribution by undersampling the dataset.}
\begin{subtable}{0.45\textwidth}
\centering
\caption{Saliency Product $P$ and Upperbound $U$. Trained on Cityscapes $\rightarrow$ ACDC and tested on ACDC Val. While $P=2^8$ and $U=1.00$ are optimal, our model requires minimal engineering due to the small mIoU range (1.7\%). }
\resizebox{0.55\textwidth}{!}{\begin{tabular}{c|ccc}
\toprule
\textbf{Parameters}    & \textbf{aAcc}  & \textbf{mIoU}  & \textbf{mAcc}  \\ \hline
P=$2^7$, U=1.00    & 84.7 & 60.1 & 73.2 \\ 
P=$2^9$, U=1.00    & 84.2  & 61.4  & \textbf{74.1}  \\
P=$2^8$, U=0.50  & 85.2  & 60.1  & 72.5  \\
P=$2^8$, U=0.75 & 82.4  & 60.3  & 73.1  \\
P=$2^8$, U=1.00    & \textbf{85.3}  & \textbf{61.8}  & 73.8  \\ \toprule
\end{tabular}
}
\label{tab:pro_bound}
\end{subtable}
\hspace{0.35in}
\begin{subtable}{0.45\textwidth}
\centering
\caption{Balanced Undersampling Strategy. Trained on Cityscapes and tested on Cityscapes Val using Supervised strategy in Section~\ref{subsec:da}. Undersampling images negatively impacts segmentation performances.}
\resizebox{0.8\textwidth}{!}{\begin{tabular}{c|c|c|ccc}
\toprule
\textbf{Sample}          & \textbf{Percent}       & \textbf{Method} & \textbf{aAcc} & \textbf{mIoU} & \textbf{mAcc} \\ \hline
\multirow{4}{*}{Uniform} & \multirow{2}{*}{25\%}  & SegFormer        & 95.0          & 71.0          & 79.1          \\  
                         &                        & Ours             & 95.1          & 71.9          & 80.0          \\ \cline{2-6} 
                         & \multirow{2}{*}{50\%}  & SegFormer        & 95.3          & 73.6          & 81.4          \\  
                         &                        & Ours             & 95.5          & 74.4          & 82.1          \\ \hline
\multirow{2}{*}{None}       & \multirow{2}{*}{100\%} & SegFormer        & 95.6          & 75.3          & 82.9          \\ 
                         &                        & Ours             & \textbf{95.9} & \textbf{76.8} & \textbf{84.2} \\ \toprule
\end{tabular}
}
\vspace{-0.3in}
\label{tab:undersampling}
\end{subtable}
\end{table*}

\subsection{Domain Adaptive Object Detection}
\label{subsec:daod-results}

\noindent
Tables~\ref{tab:bdd_dense_foggy},~\ref{tab:bdd100k_clear2rainy},~\ref{tab:bdd100k_day2night} and~\ref{tab:bdd100k_acdc} show our method improves the state-of-the-art 2PCNet~\cite{kennerley20232pcnet} across lighting, weather and geographic adaptations. Besides, our instance-level saliency guidance consistently outperforms approximate saliency guidance~\cite{thavamani2021fovea, ghosh2023learned}. Note that 2PCNet~\cite{kennerley20232pcnet} focused solely on Day $\rightarrow$ Night adaptation with its NightAug augmentation, which we only adopted for BDD100K Day $\rightarrow$ Night (Table~\ref{tab:bdd100k_day2night}). 


Figure~\ref{fig:det_day_night} shows examples where 2PCNet~\cite{kennerley20232pcnet} mistakes parts of the roads as cars and streetlights as traffic lights (For other domain adaptations, see \pointToAppendix{\ref{appendix:addl_analysis}}{E}). In contrast, our method accurately distinguish foreground objects from background elements in challenging lighting and weather.



\subsection{Domain Adaptive Semantic Segmentation}
\label{subsec:dass-results}

\noindent
Our approach is task-agnostic, allowing integration into DAFormer~\cite{hoyer2022daformer}, HRDA~\cite{hoyer2022hrda} and MIC~\cite{hoyer2023mic} for domain adaptive semantic segmentation. We observe significant improvements on ACDC and DarkZurich dataset (Table~\ref{tab:cs2acdc} and Table~\ref{tab:cs2dz}) via integration our method into all these algorithms. Besides, our instance-level image saliency guidance outperforms Static Prior~\cite{thavamani2021fovea} and Geometric Prior~\cite{ghosh2023learned}. \textbf{For comparisons with HRDA and MIC, see \pointToAppendix{\ref{appendix:mic-hrda}}{B}}. Note that HRDA and MIC use heavy resize and multi-crop augmentations during training and sliding window inference at test, significantly increasing training memory (+5 GB) and inference latency (+800 ms on an RTX 4090 GPU). In contrast, our approach requires minimal additional training memory (+0.04 GB), negligible training latency (+5.7 ms on an RTX 4090 GPU) and no additional inference latency.

Tables~\ref{tab:cs2acdc} and~\ref{tab:cs2dz} also demonstrates our method's superiority across different scenarios (w/ and w/o UDA), indicating that the learnt backbone features generalize better to unseen domains, since the backbone model already improves after supervised pre-training with image warping solely on source. A visual diagnosis of the learned backbones can be found in \pointToAppendix{\ref{appendix:addl_analysis}}{E}, including GradCam~\cite{selvaraju2017grad} visualizations showing higher discriminability and an increased focus on salient objects when saliency guided warping is employed while training. Additionally, Figure~\ref{fig:seg_acdc} illustrates the superior performance of our method in identifying both background (e.g., sky, sidewalks, buildings) and objects (e.g., traffic lights, traffic signs).

\subsection{Ablation Study}
\label{subsec:ablation}


\noindent
For all the ablations below, we use Cityscape $\rightarrow$ ACDC adaptation, except for `Distribution Balancing by Undersampling', where we fully supervised on Cityscapes. 

\noindent \textbf{Saliency Product $P$ and Upperbound $U$.} Varying $P$ and $U$ shows optimal performance at $P=2^8$ and $U=1.0$, according to quantitative scores in Table~\ref{tab:pro_bound}. Hence, $P=2^8$ and $U=1.0$ are used for all major experiments.

\noindent\textbf{Distribution Balancing by Undersampling.} We analyze undersampling strategies that select images to create a more uniform object-background distribution. Table~\ref{tab:undersampling} shows models trained with 25\% and 50\% undersampling on the Cityscapes dataset perform worse than those trained on the complete dataset, highlighting the ineffectiveness of simple undersampling to address object-background imbalance.

\noindent\textbf{Warping on Source vs. Target.} Warping in the target domain for unsupervised domain adaptation is infeasible for instance-level saliency guidance because it requires groundtruth labels. However, Table~\ref{tab:warp_bboxes} shows that warping source domain images using our instance-level saliency guidance outperforms using approximated saliency~\cite{thavamani2021fovea,ghosh2023learned} with both source and target domain warping. We also experimented with warping during testing, but this approach introduced additional latency and failed to improve performance compared to warping only the source.


\noindent \textbf{Warping with From-Seg vs. Groundtruth bboxes.}  For object detection, we utilize ground truth bounding boxes for saliency guidance. However, semantic segmentation tasks often lack this data, leading us to use bounding boxes computed from segmentations, `from-seg' for short. Results indicate superior performance with ground truth boxes, but `from-seg' boxes are a close second, highlighting their effectiveness when ground truth is unavailable (Table~\ref{tab:warp_bboxes}).

\begin{table}[t]
\centering
\small
\caption{\textbf{Ablation: Warping (source/target domain) and Bounding Box Type (from-seg/ground truth)}. Trained on Cityscapes $\rightarrow$ ACDC. Tested on ACDC Val. Using ground truth bounding boxes outperforms using bounding boxes derived from semantic segmentation. For Static prior and Geometric prior, we can apply warping during target adaptation and testing as they do not rely on ground truth. For Inst., \textit{target and test warping is impossible as ground truth boxes are unavailable in the target domain}, as per the definition of unsupervised domain adaptation.}
\label{table:src-tgt-ablation}
\resizebox{\columnwidth}{!}{%
\begin{tabular}{l|l|c|ccc}
\toprule
\multicolumn{1}{c|}{\textbf{Method}}       & \multicolumn{1}{c|}{\textbf{Warp}} & \textbf{Bboxes} & \textbf{aAcc} & \textbf{mIoU} & \textbf{mAcc} \\ \hline
DAFormer               & -             & -               & 81.9          & 57.6          & 70.6          \\ \hline
\multirow{3}{*}{+ Ours w/ Sta. Prior}  
& src \& tgt  \& test    & -               &  82.5        &    55.3      & 68.3  \\

& src \& tgt    & -               & 83.8          & 57.4          & 69.6          \\ 
                       & src           & -               & 82.3          & 59.1          & 73.3          \\ \hline
 \multirow{3}{*}{+ Ours w/ Geo. Prior}                    
  & src \& tgt  \& test    & -               &  82.2        &   57.7       &  71.6  \\
 & src \& tgt    & -               & 81.9          & 56.5          & 70.9          \\ 
   & src           & -               & 83.2          & 58.4          & 70.9          \\  \hline 
\multirow{2}{*}{\vtop{\hbox{\strut + Ours w/ Inst.}\hbox{\strut (tgt \& test inapplicable)}}}  & src           & from-seg          & 84.0          & 60.1          & 73.5          \\
                       & src           & gt              & \textbf{85.3} & \textbf{61.8} & \textbf{73.8} \\ \toprule
\end{tabular}%
}
\vspace{-0.15in}
\label{tab:warp_bboxes}
\end{table}

\section{Limitations and Conclusion}
\noindent
This work demonstrates that saliency-guided in-place image warping enhances Unsupervised Domain Adaptation (UDA). Unlike prior approaches that uniformly process objects and backgrounds, our instance-level saliency guidance oversamples object regions and undersamples backgrounds, improving backbone feature learning and aiding domain adaptation. Extensive experiments validate the effectiveness and efficiency of our approach, which adds minimal training overhead, no inference latency, and is agnostic to the adaptation task, target domain, saliency guidance, model architecture, and domain adaptation algorithm.


A limitation of our method is its reduced effectiveness in densely populated scenes where there is limited background to reduce for enlarging foreground objects. Our in-place warping strategy is less effective when background is less than one-fourth of the image. Besides, our approach is less effective on synthetic datasets like Synthia~\cite{ros2016synthia}. However, it proves effective on other synthetic datasets like GTA5~\cite{richter2016playing}, which feature \textit{realistic} city driving scenes. 

\noindent \textbf{Acknowledgment:} This work was supported by a research contract from General Motors Research-Israel, NSF Grant CNS-2038612, a US DOT grant 69A3551747111 through the Mobility21 UTC and grants 69A3552344811 and 69A3552348316 through the Safety21 UTC.



{\small
\bibliographystyle{ieee_fullname}
\bibliography{egbib}
}


\clearpage

\appendix

\section{Synthetic to Real Domain Adaptation}
\label{appendix:sim2real}

\begin{figure*}[t!]
    \centering
    \includegraphics[width=\linewidth]{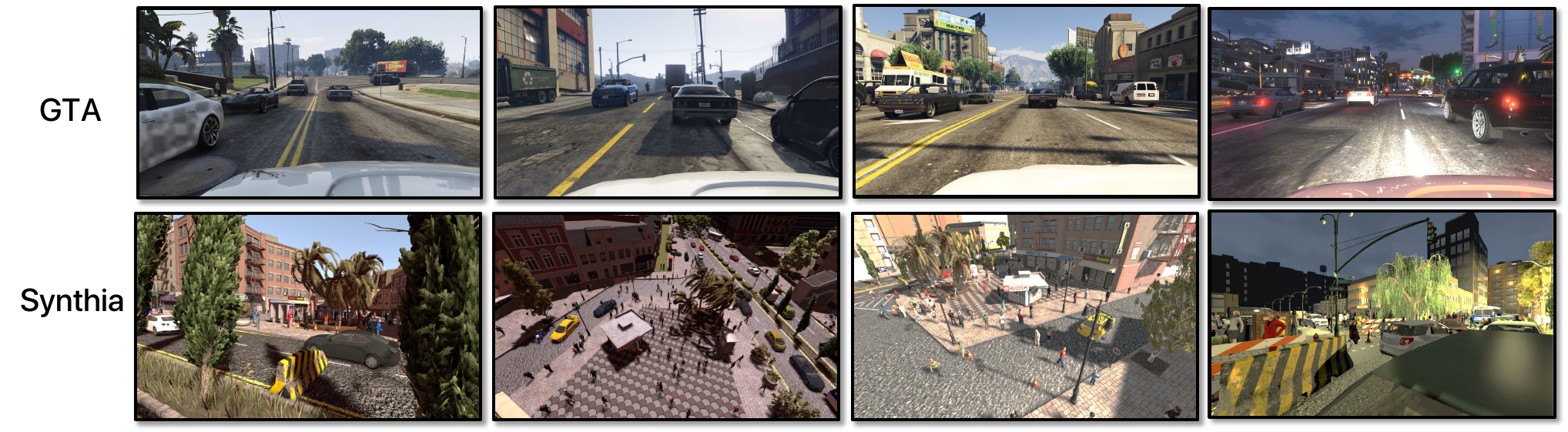}
    \caption{\textbf{Synthetic to Real Domain Adaptation}. GTA~\cite{richter2016playing} consists of realistic scenes from a vehicle's perspective with accurate road traffic simulation, closely resembling real-world driving conditions. In contrast, Synthia~\cite{ros2016synthia} features unrealistic virtual-world images that differ significantly from natural driving scenes.}
    \label{fig:gta_synthia}
\end{figure*}

\noindent
The \textit{Synthetic-to-Real (Sim2Real) Domain Gap is not the focus of this work, as synthetic scene datasets may have suboptimal simulations and inconsistent viewpoints, resulting in unrealistic object-background distributions with non-dominant backgrounds, where our strategy of oversampling objects is less effective.}

However, we do explore how our method impacts performance on the Sim2Real domain gap by considering the GTA~\cite{richter2016playing} and Synthia~\cite{ros2016synthia} synthetic datasets. We expect our method to perform better with realistic datasets and less effectively with unrealistic datasets.

As shown in Fig. \ref{fig:gta_synthia}, Grand Theft Auto V (GTA) is a role-playing game with a city driving component and \textit{realistic} traffic, making images closely resemble real driving scenes. In contrast, Synthia is a rendered dataset of a virtual city with \textit{unrealistic} traffic simulation of dynamic objects like cars and people. Additionally, many Synthia images are not captured from a vehicle's perspective, making them less representative of real driving scenes. 

Table~\ref{tab:gta_cs} shows that our method significantly improves domain adaptation from GTA $\rightarrow$ Cityscapes, while Table~\ref{tab:synthia_cs} shows only marginal improvement for Synthia $\rightarrow$ Cityscapes. This result aligns with our expectations for both source datasets.

\begin{table*}[t]
\centering
\caption{\textbf{Synthetic to Real Domain Adaptation: GTA $\rightarrow$ Cityscapes Semantic Segmentation.} Tested on Cityscapes Val. We were unable to reproduce DAFormer's~\cite{hoyer2022daformer} results averaged over 3 random seeds. Therefore, we present results with seed=0 for both DAFormer~\cite{hoyer2022daformer} and our method (DAFormer with instance-level saliency guidance). Our method shows improvement in many individual categories for GTA. Although our focus is on real datasets rather than synthetic datasets, GTA \textit{closely resembles natural driving scenes due to its realistic traffic simulation and vehicle-perspective imagery} (Figure~\ref{fig:gta_synthia}). These \textit{realistic} object-background appearances with background dominance contribute to the effectiveness of our method.}
\resizebox{\textwidth}{!}{%
\begin{tabular}{l|c|ccccccccccccccccccc}
\toprule
Method               & \textbf{mIoU}          & road          & \begin{tabular}[c]{@{}c@{}}side\\ walk\end{tabular} & building      & wall          & fence         & pole          & \begin{tabular}[c]{@{}c@{}}traffic\\ light\end{tabular} & \begin{tabular}[c]{@{}c@{}}traffic\\ sign\end{tabular} & \begin{tabular}[c]{@{}c@{}}vege-\\ tation\end{tabular} & terrain       & sky           & person        & rider         & car           & truck         & bus           & train         & \begin{tabular}[c]{@{}c@{}}motor\\ cycle\end{tabular} & bike       \\ \midrule
CBST~\cite{zou2018unsupervised}                          & 45.9          & 91.8          & 53.5                                                & 80.5          & 32.7          & 21.0          & 34.0          & 28.9                                                    & 20.4                                                   & 83.9                                                   & 34.2          & 80.9          & 53.1          & 24.0          & 82.7          & 30.3          & 35.9          & 16.0          & 25.9                                                  & 42.8          \\ 
DACS~\cite{tranheden2021dacs}                          & 52.1          & 89.9          & 39.7                                                & 87.9          & 30.7          & 39.5          & 38.5          & 46.4                                                    & 52.8                                                   & 88.0                                                   & 44.0          & 88.8          & 67.2          & 35.8          & 84.5          & 45.7          & 50.2          & 0.0           & 27.3                                                  & 34.0          \\ 
CorDA~\cite{wang2021domain}                        & 56.6          & 94.7          & 63.1                                                & 87.6          & 30.7          & 40.6          & 40.2          & 47.8                                                    & 51.6                                                   & 87.6                                                   & 47.0          & 89.7          & 66.7          & 35.9          & 90.2          & 48.9          & 57.5          & 0.0           & 39.8                                                  & 56.0          \\
ProDA~\cite{zhang2021prototypical}                        & 57.5          & 87.8          & 56.0                                                & 79.7          & 46.3          & 44.8          & 45.6          & 53.5                                                    & 53.5                                                   & 88.6                                                   & 45.2          & 82.1          & 70.7          & 39.2          & 88.8          & 45.5          & 59.4          & 1.0           & 48.9                                                  & 56.4          \\
DAFormer~\cite{hoyer2022daformer}                     & 68.3          & \textbf{95.7} & \textbf{70.2}                                       & 89.4          & 53.5          & 48.1          & \textbf{49.6} & 55.8                                                    & 59.4                                                   & \textbf{89.9}                                          & 47.9          & \textbf{92.5} & 72.2          & 44.7          & 92.3          & 74.5          & 78.2          & 65.1          & 55.9                                                  & 61.8          \\ \midrule \midrule
DAFormer (seed=0) & 66.9          & 92.6          & 58.9                                                & 89.3          & 54.2          & 42.7          & 49.4          & 57.0                                                    & 55.8                                                   & 89.2                                                   & 49.8          & 89.5          & \textbf{72.7} & 41.7          & 92.0          & 62.0          & \textbf{82.8} & \textbf{71.3} & 56.5                                                  & 62.9          \\ 
+ Ours                 & \textbf{68.5} & 92.9          & 60.0                                                & \textbf{89.8} & \textbf{55.9} & \textbf{51.5} & 49.0          & \textbf{57.2}                                           & \textbf{62.2}                                          & 89.6                                                   & \textbf{50.2} & 91.5          & 71.9          & \textbf{44.8} & \textbf{93.0} & \textbf{78.7} & 79.8          & 63.6          & \textbf{56.6}                                         & \textbf{63.6} \\ \toprule
\end{tabular}%
}
\label{tab:gta_cs}
\end{table*}

\begin{table*}[]
\centering
\scriptsize
\caption{\textbf{Synthetic to Real Domain Adaptation: Synthia $\rightarrow$ Cityscapes Semantic Segmentation.} Tested on Cityscapes Val. We were unable to reproduce DAFormer's~\cite{hoyer2022daformer} results averaged over three random seeds. Therefore, we present results with seed=0 for both DAFormer~\cite{hoyer2022daformer} and our method (DAFormer with instance-level saliency guidance). Our improvement on Synthia $\rightarrow$ Cityscapes is marginal, which we attribute to the fact that \textit{Synthia does not exhibit appearance akin to natural driving scenes} (Figure~\ref{fig:gta_synthia}). These \textit{unrealistic} object-background appearances make our method ineffective. }
\resizebox{\textwidth}{!}{%
\begin{tabular}{l|c|ccccccccccccccccccc}
\toprule
Method               & \textbf{mIoU}          & road          & \begin{tabular}[c]{@{}c@{}}side\\ walk\end{tabular} & building      & wall          & fence        & pole          & \begin{tabular}[c]{@{}c@{}}traffic\\ light\end{tabular} & \begin{tabular}[c]{@{}c@{}}traffic\\ sign\end{tabular} & \begin{tabular}[c]{@{}c@{}}vege-\\ tation\end{tabular} & terrain & sky           & person        & rider         & car           & truck & bus           & train & \begin{tabular}[c]{@{}c@{}}motor\\ cycle\end{tabular} & bike       \\ \midrule
CBST~\cite{zou2018unsupervised}                          & 42.6          & 68.0          & 29.9                                                & 76.3          & 10.8          & 1.4          & 33.9          & 22.8                                                    & 29.5                                                   & 77.6                                                   & -       & 78.3          & 60.6          & 28.3          & 81.6          & -     & 23.5          & -     & 18.8                                                  & 39.8          \\ 
DACS~\cite{tranheden2021dacs}                          & 48.3          & 80.6          & 25.1                                                & 81.9          & 21.5          & 2.9          & 37.2          & 22.7                                                    & 24.0                                                   & 83.7                                                   & -       & \textbf{90.8} & 67.6          & 38.3          & 82.9          & -     & 38.9          & -     & 28.5                                                  & 47.6          \\ 
CorDA~\cite{wang2021domain}                         & 55.0          & \textbf{93.3} & \textbf{61.6}                                       & 85.3          & 19.6          & 5.1          & 37.8          & 36.6                                                    & 42.8                                                   & 84.9                                                   & -       & 90.4          & 69.7          & 41.8          & 85.6          & -     & 38.4          & -     & 32.6                                                  & 53.9          \\ 
ProDA~\cite{zhang2021prototypical}                         & 55.5          & 87.8          & 45.7                                                & 84.6          & 37.1          & 0.6          & 44.0          & 54.6                                                    & 37.0                                                   & \textbf{88.1}                                          & -       & 84.4          & \textbf{74.2} & 24.3          & 88.2          & -     & 51.1          & -     & 40.5                                                  & 45.6          \\ 
DAFormer~\cite{hoyer2022daformer}                      & \textbf{60.9} & 84.5          & 40.7                                                & \textbf{88.4} & 41.5          & 6.5          & \textbf{50.0} & \textbf{55.0}                                           & \textbf{54.6}                                          & 86.0                                                   & -       & 89.8          & 73.2          & \textbf{48.2} & 87.2          & -     & \textbf{53.2} & -     & 53.9                                                  & 61.7          \\  \midrule \midrule
DAFormer (seed=0) & 59.2          & 87.8          & 47.6                                                & 87.6          & 43.7          & 5.8          & 49.0          & 48.0                                                    & 53.1                                                   & 82.3                                                   & -       & 71.8          & 71.7          & 46.0          & 87.5          & -     & 49.3          & -     & 52.7                                                  & \textbf{63.4} \\ 
+ Ours                 & 59.4          & 92.7          & 57.7                                                & 87.1          & \textbf{44.8} & \textbf{7.1} & 46.2          & 46.9                                                    & 53.6                                                   & 80.6                                                   & -       & 68.2          & 72.5          & 42.5          & \textbf{90.6} & -     & 41.6          & -     & \textbf{55.2}                                         & 63.3          \\ \toprule
\end{tabular}%
}
\label{tab:synthia_cs}
\end{table*}

\clearpage
\section{Comparison with other Domain Adaptation Strategies}
\label{appendix:mic-hrda}

\begin{table}[]
\centering
\small
\caption{\textbf{Comparison with other Domain Adaptation Strategies: Computational Efficiency.} We use a single RTX 4090 GPU with a batch size of 1, maintaining the same training and inference image size of 512x1024 as specified in DAFormer~\cite{hoyer2022daformer}. We also include a comparison using a single A40 GPU, which has more memory, enabling us to experiment with an image size of 768x1536.} Our saliency-guided image warping imposes minimal memory overhead during training. Additionally, since we do not perform warping during test time, our method incurs no inference latency overhead.
\resizebox{\columnwidth}{!}{%
\begin{tabular}{c|l|c|r}
\toprule
\textbf{Image Size} &  \textbf{Method} & \textbf{Training Memory} & \textbf{Inference Time} \\  \hline

 \multirow{3}{*}{512x1024}  & DAFormer      & 16.41 GB  &  195.0 ms     \\ 
   & \textbf{DAFormer + Ours}  & \textbf{+ 0.04 GB}  & \textbf{+ 0 ms}    \\ 
  & DAFormer + HRDA  & + 5.11 GB  &  + 801.2 ms   \\ \midrule

 \multirow{3}{*}{768x1536}  & DAFormer      & 30.40 GB   &   261.3 ms    \\ 
   & \textbf{DAFormer + Ours}  &  \textbf{+ 0.04 GB} & \textbf{+ 0 ms}    \\ 
  & DAFormer + HRDA  & + 6.69 GB   &    +1074.0 ms \\ \midrule  
  
\end{tabular}%
}
\label{tab:memory}
\end{table}


\subsection{Differences between HRDA/MIC and DAFormer} 

\noindent
Domain adaptation extensions have been proposed for the DAFormer~\cite{hoyer2022daformer} training framework, such as HRDA~\cite{hoyer2022hrda} and MIC~\cite{hoyer2023mic}, both of which demonstrate strong performance. HRDA~\cite{hoyer2022hrda} introduces a multi-scale high-resolution crop training strategy combined with sliding window inference, whereas MIC~\cite{hoyer2023mic} implements a masking consistency strategy for target domain images to enhance spatial context.

\subsection{Computational Efficiency} 

\noindent
Table~\ref{tab:memory} shows the computational efficiency comparison using the DAFormer image scale. We observe that incorporating HRDA into DAFormer significantly increases training and inference computational costs due to HRDA's train-time HR-cropping and test-time sliding window inference. In contrast, our inference costs \textit{remain the same} as those of DAFormer~\cite{hoyer2022daformer}, \textit{as we do not warp during test time}. Our training memory consumption are only slightly higher, due to the lightweight design of our warping-related modules. As we noted in the main manuscript, no additional learned parameters are introduced. Moreover, the latency of the warping and unwarping operations is minimal, at 1.5 ms and 4.2 ms respectively.



\subsection{Evaluation Methodology} 

\noindent
Based on the above discussion, we apply HRDA's and MIC's strategies to DAFormer. However, for a fair comparison, we perform training and inference \textit{following DAFormer's~\cite{hoyer2022daformer} training/testing image scales and evaluation paradigm}. An alternative comparison would involve training and evaluating on full-scale images to match HRDA, which deviates from original DAFormer~\cite{hoyer2022daformer} training and evaluation that uses half-scaled images. Unfortunately, we lack access to a GPU with high memory capacity, such as Tesla A100 with 80 GB of memory.

\subsection{Results and Analysis} 

\noindent \textbf{Cityscapes $\rightarrow$ DarkZurich/ACDC Semantic Segmentation:}  Table~\ref{tab:cs_dz_2} and~\ref{tab:cs_acdc_2} present results for Cityscapes $\rightarrow$ DarkZurich and Cityscapes $\rightarrow$ ACDC semantic segmentation, respectively. MIC (HRDA) refers to MIC and HRDA training strategies added to DAFormer. We compare DAFormer and MIC (HRDA) with and without our instance-level saliency guided image warping. We observe smaller improvements with our method when combined with MIC (HRDA) due to HRDA's use of multi-scale cropping. See next paragraph for explanations.


\noindent \textbf{Ablation on Cityscapes $\rightarrow$ ACDC Semantic Segmentation:} We studied the interaction of our saliency-guided image warping with MIC~\cite{hoyer2023mic} and HRDA~\cite{hoyer2022hrda} individually, as our method is orthogonal and plug-and-play. Table~\ref{tab:michrdaablation} shows that \textit{(a)} HRDA did not improve DAFormer when trained at DAFormer's scale (\textit{1024 $\times$ 512}) instead of full scale (\textit{2048 $\times$ 1024}). \textit{(b)} Our method improve performance for MIC and HRDA, but neither combination outperform DAFormer+Ours. We investigate and find that our limited improvement on HRDA is due to their use of `detail crop'. \textit{The HRDA detail crop focuses on small object regions, which our warping already oversamples. This results in redundant efforts and reduce overall effectiveness.} To verify this, we remove the detail crop from HRDA and observe that our method showed better improvement on this HRDA$^*$ variant (see Table~\ref{tab:michrdaablation}).


\begin{table*}[t]
\centering
\caption{\textbf{Comparison with other Domain Adaptation Strategies: Cityscapes $\rightarrow$ DarkZurich Semantic Segmentation.} Tested on DarkZurich Val. Our method improves IoU scores in conjunction with both DAFormer and MIC (HRDA) strategies. We observe smaller improvements with MIC (HRDA) because both our warping and HRDA's detail-crop focus on small object regions, leading to redundant efforts. }
\resizebox{\textwidth}{!}{%
\begin{tabular}{l|c|ccccccccccccccccccc}
\toprule
Method   & \textbf{mIoU}          & road          & \begin{tabular}[c]{@{}c@{}}side\\ walk\end{tabular} & building      & wall          & fence         & pole          & \begin{tabular}[c]{@{}c@{}}traffic\\ light\end{tabular} & \begin{tabular}[c]{@{}c@{}}traffic\\ sign\end{tabular} & \begin{tabular}[c]{@{}c@{}}vege-\\ tation\end{tabular} & terrain       & sky           & person        & rider         & car           & truck & bus & train & \begin{tabular}[c]{@{}c@{}}motor\\ cycle\end{tabular} & bike          \\  \midrule 
DAFormer          & 33.8          & \textbf{92.0} & 66.7                                                & 47.3          & 25.9          & \textbf{50.8} & 38.0          & 24.6                                                    & 19.4                                                   & \textbf{62.2}                                          & 31.9          & 17.9          & 20.4          & \textbf{28.4} & 61.7          & -     & -   & -     & 21.0                                                  & \textbf{34.3} \\
DAFormer + Ours   & \textbf{37.1} & 88.7          & \textbf{70.9}                                       & \textbf{60.1} & \textbf{42.1} & 49.6          & \textbf{39.8} & \textbf{47.9}                                           & \textbf{21.5}                                          & 49.8                                                   & \textbf{35.7} & \textbf{25.4} & \textbf{23.8} & 25.9          & \textbf{66.8} & -     & -   & -     & \textbf{24.3}                                         & 32.4          \\ \midrule \midrule
MIC (HRDA)        & 39.8          & \textbf{78.8} & \textbf{13.0}                                       & 83.1          & 46.7          & 52.2          & 42.2          & \textbf{44.5}                                           & 28.8                                                   & 64.3                                                   & 35.4          & \textbf{82.8} & 24.9          & \textbf{33.8} & 62.8          & -     & -   & -     & \textbf{25.9}                                         & \textbf{37.3} \\
MIC (HRDA) + Ours & \textbf{40.1} & 75.8          & 5.8                                                 & \textbf{83.3} & \textbf{55.7} & \textbf{54.8} & \textbf{44.3} & 30.5                                                    & \textbf{38.0}                                          & \textbf{66.3}                                          & \textbf{42.7} & 82.4          & \textbf{39.4} & 15.1          & \textbf{69.3} & -     & -   & -     & 25.1                                                  & 32.9  \\ \toprule       
\end{tabular}%
}
\label{tab:cs_dz_2}
\end{table*}

\begin{table*}[t]
\centering
\caption{\textbf{Comparison with other Domain Adaptation Strategies: Cityscapes $\rightarrow$ ACDC Semantic Segmentation.} Tested on ACDC Val. Our method improves IoU score in conjunction with both DAFormer and MIC (HRDA) strategies. We observe smaller improvements with MIC (HRDA) because both our warping and HRDA's detail-crop focus on small object regions, leading to redundant efforts. }
\resizebox{\textwidth}{!}{%
\begin{tabular}{l|c|ccccccccccccccccccc}
\toprule
Method   & \textbf{mIoU}          & road          & \begin{tabular}[c]{@{}c@{}}side\\ walk\end{tabular} & building      & wall          & fence         & pole          & \begin{tabular}[c]{@{}c@{}}traffic\\ light\end{tabular} & \begin{tabular}[c]{@{}c@{}}traffic\\ sign\end{tabular} & \begin{tabular}[c]{@{}c@{}}vege-\\ tation\end{tabular} & terrain       & sky           & person        & rider         & car           & truck         & bus           & train         & \begin{tabular}[c]{@{}c@{}}motor\\ cycle\end{tabular} & bike          \\ \midrule 
DAFormer          & 57.6          & 72.7          & \textbf{57.5}                                       & \textbf{80.1} & 42.5          & 38.0          & 50.9          & 45.1                                                    & 50.0                                                   & 71.1                                                   & \textbf{38.5} & 67.0          & 56.0          & 29.9          & 81.8          & 76.6          & 78.9          & 79.9          & 40.8                                                  & 36.7          \\
DAFormer + Ours   & \textbf{61.8} & \textbf{82.9} & 56.1                                                & 79.8          & \textbf{44.6} & \textbf{40.3} & \textbf{52.7} & \textbf{60.8}                                           & \textbf{52.5}                                          & \textbf{72.0}                                          & 38.4          & \textbf{78.0} & \textbf{56.6} & \textbf{30.5} & \textbf{84.9} & \textbf{80.2} & \textbf{86.9} & \textbf{86.4} & \textbf{44.5}                                         & \textbf{45.8} \\ \midrule \midrule
MIC (HRDA)        & 59.6          & 66.4          & \textbf{55.8}                                       & \textbf{84.8} & \textbf{54.6} & \textbf{41.3} & 49.7          & 41.2                                                    & \textbf{53.6}                                          & \textbf{78.3}                                          & 37.7          & 67.4          & 58.3          & 24.0          & 83.6          & 70.1          & 88.8          & 87.9          & \textbf{40.7}                                         & 48.4          \\
MIC (HRDA) + Ours & \textbf{61.6} & \textbf{80.6} & 38.6                                                & 83.7          & 50.5          & \textbf{41.3} & \textbf{50.5} & \textbf{56.7}                                           & 49.6                                                   & 75.2                                                   & \textbf{39.4} & \textbf{83.9} & \textbf{58.9} & \textbf{32.0} & \textbf{86.0} & \textbf{75.4} & \textbf{91.8} & \textbf{88.0} & 40.2                                                  & \textbf{48.5} \\ \toprule
\end{tabular}%
}
\label{tab:cs_acdc_2}
\end{table*}



\begin{table*}[t]
\centering
\caption{\textbf{Comparison with other Domain Adaptation Strategies: Ablation on Cityscapes $\rightarrow$ ACDC Semantic Segmentation.} Tested on ACDC Val. Our method improves performance when combined with base DAFormer~\cite{hoyer2022daformer}, MIC~\cite{hoyer2023mic}, HRDA~\cite{hoyer2022hrda}, and HRDA without HR-detail crop (HRDA$^*$). Our observations are -- (a) Instance-level saliency guidance enhances MIC but does not exceed the performance of DAFormer with instance-level saliency guidance. (b) HRDA performs worse than DAFormer at DAFormer's training scales, indicating the necessity for full-resolution training for HRDA to perform well, consistent with findings in~\cite{hoyer2022hrda}. -- (c) Removing the HR-detail crop (HRDA$^*$) allows adding our warping method to achieve greater performance improvements. This is because both our warping and HRDA’s detail crop focus on small object regions, resulting in redundant efforts.}
\resizebox{\textwidth}{!}{%
\begin{tabular}{l|c|ccccccccccccccccccc}
\toprule
Method                     & \textbf{mIoU}          & road          & \begin{tabular}[c]{@{}c@{}}side\\ walk\end{tabular} & building      & wall          & fence         & pole          & \begin{tabular}[c]{@{}c@{}}traffic\\ light\end{tabular} & \begin{tabular}[c]{@{}c@{}}traffic\\ sign\end{tabular} & \begin{tabular}[c]{@{}c@{}}vege-\\ tation\end{tabular} & terrain       & sky           & person        & rider         & car           & truck         & bus           & train         & \begin{tabular}[c]{@{}c@{}}motor\\ cycle\end{tabular} & bike          \\  \midrule 
DAFormer                            & 57.6          & 72.7          & \textbf{57.5}                                       & \textbf{80.1} & 42.5          & 38.0          & 50.9          & 45.1                                                    & 50.0                                                   & 71.1                                                   & \textbf{38.5} & 67.0          & 56.0          & 29.9          & 81.8          & 76.6          & 78.9          & 79.9          & 40.8                                                  & 36.7          \\
DAFormer + Ours                     & \textbf{61.8} & \textbf{82.9} & 56.1                                                & 79.8          & \textbf{44.6} & \textbf{40.3} & \textbf{52.7} & \textbf{60.8}                                           & \textbf{52.5}                                          & \textbf{72.0}                                          & 38.4          & \textbf{78.0} & \textbf{56.6} & \textbf{30.5} & \textbf{84.9} & \textbf{80.2} & \textbf{86.9} & \textbf{86.4} & \textbf{44.5}                                         & \textbf{45.8} \\ \midrule \midrule 
MIC (DAFormer)                     & 58.8          & 61.1          & 59.5                                                & 73.2          & \textbf{47.4} & \textbf{45.2} & 51.4          & 44.7                                                    & 48.2                                                   & \textbf{78.4}                                          & 38.1          & 51.4          & 60.4          & \textbf{41.3} & 84.3          & 78.5          & 84.3          & 78.9          & \textbf{43.9}                                         & \textbf{46.5} \\
MIC (DAFormer) + Ours               & \textbf{60.6} & \textbf{72.8} & \textbf{62.9}                                       & \textbf{73.4} & 45.1          & 36.5          & \textbf{52.9} & \textbf{49.0}                                           & \textbf{49.9}                                          & 76.9                                                   & \textbf{39.8} & \textbf{65.5} & \textbf{60.6} & 40.4          & \textbf{85.2} & \textbf{80.9} & \textbf{90.5} & \textbf{87.0} & 41.0                                                  & 41.3          \\  \midrule \midrule 
HRDA (DAFormer)                     & 56.9          & 79.9          & 37.8                                                & \textbf{81.1} & \textbf{45.6} & 33.9          & 47.8          & \textbf{47.3}                                           & 47.1                                                   & 74.3                                                   & 37.1          & \textbf{84.0} & 47.7          & 17.6          & \textbf{84.2} & 69.1          & 88.2          & 75.1          & 37.6                                                  & 45.1          \\
HRDA (DAFormer)  + Ours             & \textbf{57.7} & \textbf{85.6} & \textbf{48.2}                                       & 71.7          & 41.6          & \textbf{39.4} & \textbf{50.8} & 17.7                                                    & \textbf{47.8}                                          & \textbf{75.2}                                          & \textbf{37.9} & 81.4          & \textbf{56.3} & \textbf{25.0} & 82.3          & \textbf{73.4} & \textbf{88.8} & \textbf{82.3} & \textbf{45.3}                                         & \textbf{46.7} \\  \midrule \midrule 
HRDA$^*$ (DAFormer)        & 58.3          & 68.5          & 59.4                                                & 82.8          & \textbf{50.4} & \textbf{40.8} & 50.3          & 42.4                                                    & 44.4                                                   & \textbf{77.6}                                          & 38.0          & 69.4          & \textbf{55.7} & 27.2          & 83.2          & 77.4          & 78.2          & 79.0          & 34.9                                                  & 48.4          \\
HRDA$^*$ (DAFormer) + Ours & \textbf{62.1} & \textbf{89.7} & \textbf{61.1}                                       & \textbf{83.9} & 43.4          & 39.5          & \textbf{52.7} & \textbf{43.1}                                           & \textbf{45.0}                                          & 75.6                                                   & \textbf{38.7} & \textbf{86.1} & 55.0          & \textbf{28.0} & \textbf{84.9} & \textbf{81.1} & \textbf{88.5} & \textbf{86.0} & \textbf{44.9}                                         & \textbf{53.4} \\ \toprule
\end{tabular}%
}
\label{tab:michrdaablation}
\end{table*}


\clearpage
\section{Supervised Setting}
\label{appendix:supervised}

\begin{figure*}
    \centering

    \includegraphics[width=\textwidth]{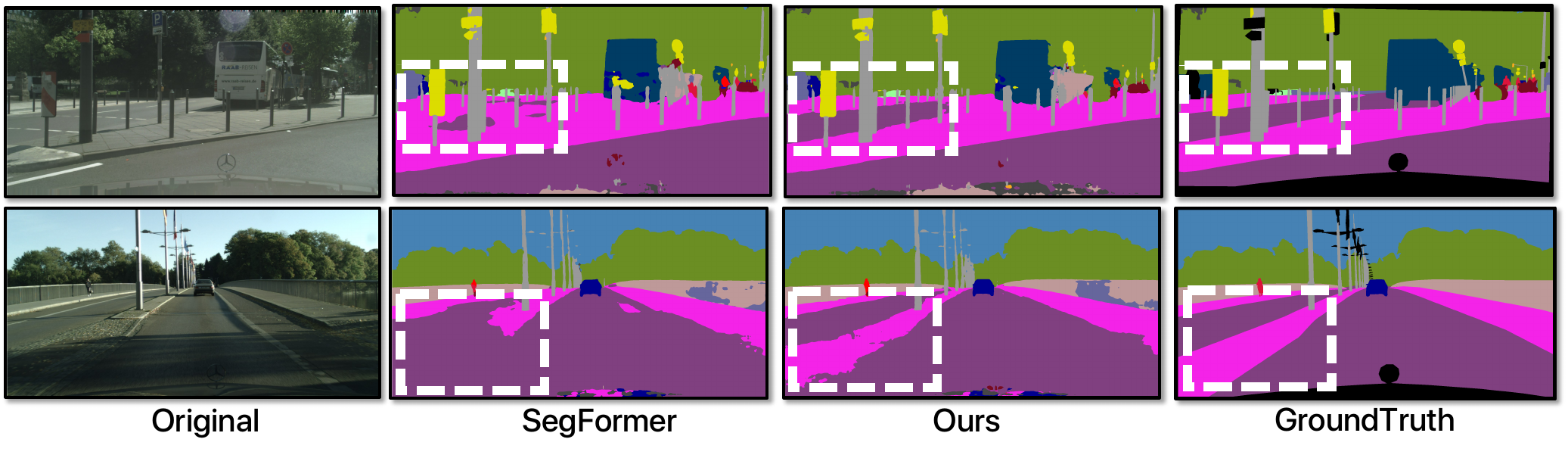}

    \caption{\textbf{Supervised Setting: Cityscapes Semantic Segmentation}. Our segmentation map more closely resembles the ground truth, indicating a more accurate understanding of objects and backgrounds in urban scenes. Notably, our method effectively distinguishes backgrounds such as sidewalks and roads, even in the presence of occlusions (top row) and shadows (bottom row). }
    \label{fig:seg_cs}
\end{figure*}

\noindent
\textit{Supervised setting is not the focus of this work. Instance-level saliency guidance is effective in domain adaptation because its focus on objects significantly reduces the negative impact of backgrounds with large cross-domain variations. In supervised settings, where background variations are low within a single-domain, this focus on objects does not provide the same benefit. } 

In Section 3.2 of the main paper, we mentioned that source pre-training improves performance in the supervised setting on the source domain. Results are presented below.

\noindent \textbf{Cityscapes Semantic Segmentation:}
We demonstrate improved semantic segmentation performance on Cityscapes by applying our method to the SegFormer model, which serves as the base architecture for the DAFormer~\cite{hoyer2022daformer} training strategy (see Table~\ref{tab:seg_cs}). Visual comparisons of our method versus SegFormer can be observed in Figure~\ref{fig:seg_cs}. 

\noindent \textbf{BDD100K Object Detection:}
We demonstrate improved object detection performance on BDD100K (Day) and BDD100 (Clear) when applying our method to Faster R-CNN, the base architecture of the 2PCNet~\cite{kennerley20232pcnet} training strategy (see Table~\ref{tab:det_day} and Table~\ref{tab:det_clear}). Note that BDD100K (Day) includes images taken during the day in both clear and bad weather, while BDD100K (Clear) includes images taken in clear weather during both day and night.

\begin{table*}[t]
\centering
\caption{\textbf{Supervised Setting: Cityscapes Semantic Segmentation}. Tested on Cityscapes Val. Instance-level saliency guided image warping improves segmentation on the source domain by \textbf{\txtgreen{+1.5 mIoU}} (along with improvements on the target domain, shown in Tables 4 and 5 in the main manuscript), indicating better learned backbone features. }
\resizebox{\textwidth}{!}{%
\begin{tabular}{l|c|ccccccccccccccccccc}
\toprule
Method          & \textbf{mIoU} & road & \begin{tabular}[c]{@{}c@{}}side\\ walk\end{tabular} & building & wall & fence & pole & \begin{tabular}[c]{@{}c@{}}traffic\\ light\end{tabular} & \begin{tabular}[c]{@{}c@{}}traffic\\ sign\end{tabular} & \begin{tabular}[c]{@{}c@{}}vege-\\ tation\end{tabular} & terrain & sky  & person & rider & car  & truck & bus  & train & \begin{tabular}[c]{@{}c@{}}motor\\ cycle\end{tabular} & bike \\ \hline

SegFormer                           & 75.3 & 98.0 & 83.4                                                & 91.8     & 59.6 & 59.6 & 57.5 & 64.0                                                    & 74.4                                                  & 91.9                                                   & 62.8    & 94.6 & 77.6   & 56.0 & 93.7 & 81.6  & 81.4 & 70.1  & 59.2                                                  & 73.3    \\  \hline
+ Ours w/ Sta. Prior      & 76.1  & 97.9 & 83.6                                               & 91.9     & 58.0 & 58.0  & 57.7 & 63.3                                                    & 73.9                                                  & 91.7                                                   & 64.4    & 94.4 & 77.7  & \textbf{57.4}  & 93.8 & 81.0  & \textbf{87.7} & 80.3  & 60.7                                                 & 72.8    \\ 
+ Ours w/ Geo. Prior                                & 76.5  & 98.0 & 84.3                                             & \textbf{92.2}    & {\textbf{61.4}} & 57.6  & 59.3 & 64.7                                                  & 74.0                                                   & 91.9                                                   & 65.3    & 94.7 & 78.2   & 55.3  & 93.9 & \textbf{83.1}  & 86.5 & 81.1  & 60.1                                                  & 72.1    \\ 
+ Ours  w/ Inst.     & \textbf{76.8} & \textbf{98.1} & \textbf{84.8}                                              & \textbf{92.2}     & 59.9 & \textbf{58.3}  & \textbf{59.6} & \textbf{65.1}                                                    & {\textbf{75.4}}                                                   & {\textbf{92.3}}                                                   & \textbf{66.2}    & \textbf{94.8} & \textbf{78.2}   & 55.3  & \textbf{94.2} & 82.0 & 85.7 & \textbf{81.3}  & \textbf{61.8}                                                 & \textbf{74.4}    \\ 
\toprule
\end{tabular}%
}
\label{tab:seg_cs}
\end{table*}

\begin{table*}[t!]
\centering
\caption{\textbf{Supervised Setting: BDD100K (Day) Object Detection.} Tested on BDD100K Day Val, which includes images captured in both good and bad weather. As shown by mAP50 (overall and per category), our saliency guided image warping improves detection performance in the source domain, and our instance-level saliency guidance is competitive compared with other saliency priors. }
\resizebox{\textwidth}{!}{%
\begin{tabular}{l|cccccc|ccccccccc}
\toprule
Method  & mAP  & \textbf{mAP50}  &  mAP75	&  mAPs	& mAPm	& mAPl & person & rider & car & truck & bus & \begin{tabular}[c]{@{}c@{}}motor\\ cycle\end{tabular} & bike & \begin{tabular}[c]{@{}c@{}}traffic\\ light\end{tabular} & \begin{tabular}[c]{@{}c@{}}traffic\\ sign\end{tabular} \\ \hline
FRCNN        & 30.1     & 56.4   &  28.1 & 13.9  &  37.6   & 51.0     & 64.0            & 50.7           & 80.3          & 62.5           & 62.9          & 45.3                                                           & 49.7             & 66.7                                                             & 69.8                                                            \\ \hline
+ Ours w/ Sta. Prior      &  30.9  & 57.1     & \textbf{28.6}   & 14.4    & \textbf{38.9}    &  \textbf{53.4}        & 65.7            & 53.1           & \textbf{80.9} & \textbf{62.7}  & 63.0          & \textbf{48.8}                                                  & 50.8             & \textbf{69.2}                                                    & 71.2                                                            \\
+ Ours w/ Geo. Prior      & \textbf{31.1}  & \textbf{57.9}  &  28.3  & \textbf{14.5}   &  38.5  & 52.7  & 66.3            & \textbf{53.6}  & 80.7          & 62.5           & 62.8          & 48.1                                                           & \textbf{52.9}    & 68.4                                                             & \textbf{71.4}                                                   \\
+ Ours w/ Inst.    &  30.7     & 57.2  &  27.9  &  \textbf{14.5}  &  38.4  &   52.8        & \textbf{66.4}   & 53.3           & 80.8          & 62.4           & \textbf{63.7} & 47.7                                                           & 51.7             & 68.6                                                             & 71.1        \\ \toprule                                                   
\end{tabular}%
}
\label{tab:det_day}
\end{table*}

\begin{table*}[t!]
\centering
\caption{\textbf{Supervised Setting: BDD100K (Clear) Object Detection.} Tested on BDD100K Clear Val, which includes both day and night images. As shown by mAP50 (overall and per category), our saliency guided image warping improves detection performance in the source domain, and our instance-level saliency guidance is competitive compared with other saliency priors.}
\resizebox{\textwidth}{!}{%
\begin{tabular}{l|cccccc|ccccccccc}
\toprule
Method & mAP  & \textbf{mAP50}  &  mAP75	&  mAPs	& mAPm	& mAPl & person & rider & car & truck & bus & \begin{tabular}[c]{@{}c@{}}motor\\ cycle\end{tabular} & bike & \begin{tabular}[c]{@{}c@{}}traffic\\ light\end{tabular} & \begin{tabular}[c]{@{}c@{}}traffic\\ sign\end{tabular} \\ \hline
FRCNN          & 25.4  & 49.6 &   22.5	& 12.2	& 30.3	& 44.2      & 59.3            & 38.8           & 76.5          & 53.2           & 54.7          & 43.1                                                           & 45.6             & 56.4                                                             & 68.7                                                            \\ \hline
+ Ours w/ Sta. Prior    & \textbf{26.0}  & 50.2  & \textbf{22.9}	& 11.9	& \textbf{31.2}	& 44.5         & 59.2            & 38.8           & \textbf{76.7} & \textbf{54.5}  & 55.7          & \textbf{45.5}                                                  & 46.0             & 56.6                                                             & 69.2                                                            \\
+ Ours w/ Geo. Prior    &  25.9 & \textbf{50.3}  & 22.8	& 12.0	& 31.0	& \textbf{44.8}  & 59.4            & \textbf{41.1}  & \textbf{76.7} & 53.7           & 56.1          & 42.7                                                           & \textbf{46.9}    & \textbf{56.7}                                                    & \textbf{69.3}                                                   \\
+ Ours w/ Inst.       &   25.9 & 50.1 &   22.8 & 	\textbf{12.2} & 	31.0	&  43.7          & \textbf{59.7}   & 37.7           & 76.6          & 54.4           & \textbf{56.9} & 43.7                                                           & 46.0             & 56.5                                                             & 69.1                                                 \\ \hline           
\end{tabular}%
}
\label{tab:det_clear}
\end{table*}

\clearpage
\section{Additional Domain Adaptations}

\begin{table*}[t]
\centering
\caption{\textbf{Additional Domain Adaptations: Cityscapes $\rightarrow$ Foggy Cityscapes.}. We observe small improvements because -- (a) The \textit{synthetic} fog introduced in Foggy Cityscapes does not accurately mimic real fog, thereby posing less of a challenge to segmentation algorithms. Consequently, the baseline model already attains high scores, making it difficult for our warping method to yield substantial improvements. -- (b) Since Foggy Cityscapes adds a fixed amount of fog to each image in Cityscapes, there is minimal cross-domain background variations. In this situation, our warping strategy that oversamples objects to mitigate negative background impacts from large cross-domain background variations is less effective. }
\resizebox{\textwidth}{!}{%
\begin{tabular}{l|c|ccccccccccccccccccc}
\toprule
Method          & \textbf{mIoU} & road & \begin{tabular}[c]{@{}c@{}}side\\ walk\end{tabular} & building & wall & fence & pole & \begin{tabular}[c]{@{}c@{}}traffic\\ light\end{tabular} & \begin{tabular}[c]{@{}c@{}}traffic\\ sign\end{tabular} & \begin{tabular}[c]{@{}c@{}}vege-\\ tation\end{tabular} & terrain & sky  & person & rider & car  & truck & bus  & train & \begin{tabular}[c]{@{}c@{}}motor\\ cycle\end{tabular} & bike \\ \hline

DAFormer                            & 74.7              & 97.8          & 83.0                                                & 88.7          & \textbf{60.4} & 59.4          & 57.4          & \textbf{60.7}                                           & 72.8                                                   & 89.2                                                   & \textbf{64.9} & 81.2          & 76.6          & 54.4          & \textbf{93.2} & 75.9          & 86.9          & 80.7          & 62.8                                                  & 73.8          \\
+ Ours w/ Inst.                           & \textbf{75.5} & \textbf{98.0} & \textbf{84.2}                                       & \textbf{89.7} & 59.7          & \textbf{61.9} & \textbf{57.7} & 60.4                                                    & \textbf{73.3}                                          & \textbf{89.4}                                          & 62.2          & \textbf{82.4} & \textbf{76.7} & \textbf{56.3} & 92.8          & \textbf{82.2} & \textbf{88.0} & \textbf{82.0} & \textbf{63.5}                                         & \textbf{73.9}  \\ \toprule
\end{tabular}%
}
\label{tab:seg_cs_fcs}
\end{table*}

\noindent
\textbf{Cityscapes $\rightarrow$ Foggy Cityscapes:} Table~\ref{tab:seg_cs_fcs} shows the result for domain adaptative object detection from Cityscapes $\rightarrow$ Foggy Cityscapes. Our improvement is small on the Foggy Cityscapes dataset. This is because \textit{(a)} the baseline is already strong when dealing with easy synthetic fog \textit{(b)} there is little cross-domain background variation, leaving minimal room for improvement with our warping approach that oversamples objects to reduce the negative impacts of cross-domain large background variations.

\clearpage
\section{Additional Analysis}
\label{appendix:addl_analysis}

\begin{figure*}[t]
   \centering
   \includegraphics[width=1\linewidth]{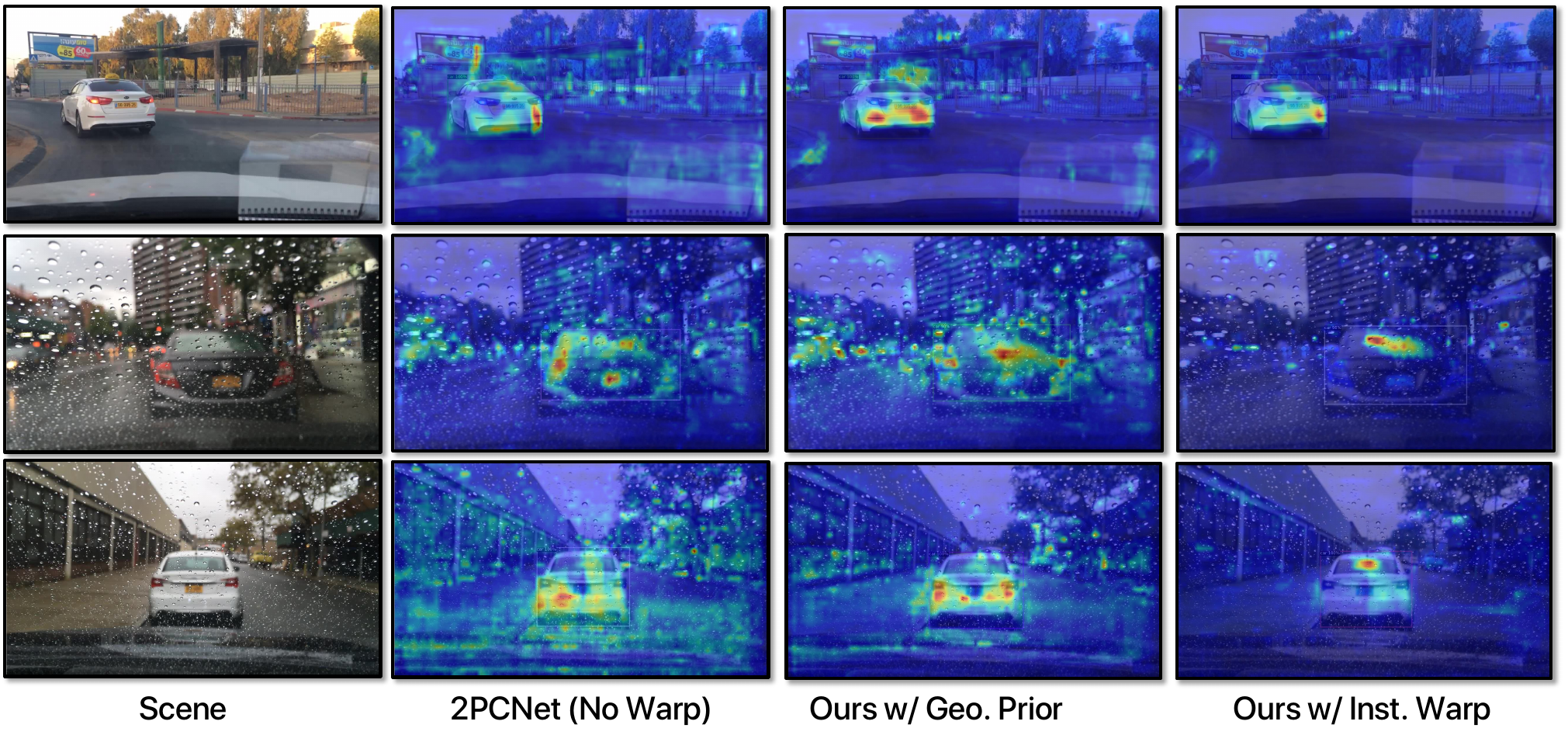} 
    \caption{\textbf{Additional Analysis: Grad-CAM~\cite{selvaraju2017grad} Visualization of the Learned Model}. We visualize the last layer feature of the learned ResNet-50 backbone. Grad-CAM visualization shows that the model trained with our method demonstrate a higher focus on salient objects, indicating better-learned features and improved scene comprehension.}
    \label{fig:grad_cam}
    \vspace{-0.1in}
\end{figure*}

\begin{figure*}[t]
   \centering
   \includegraphics[width=1\linewidth]{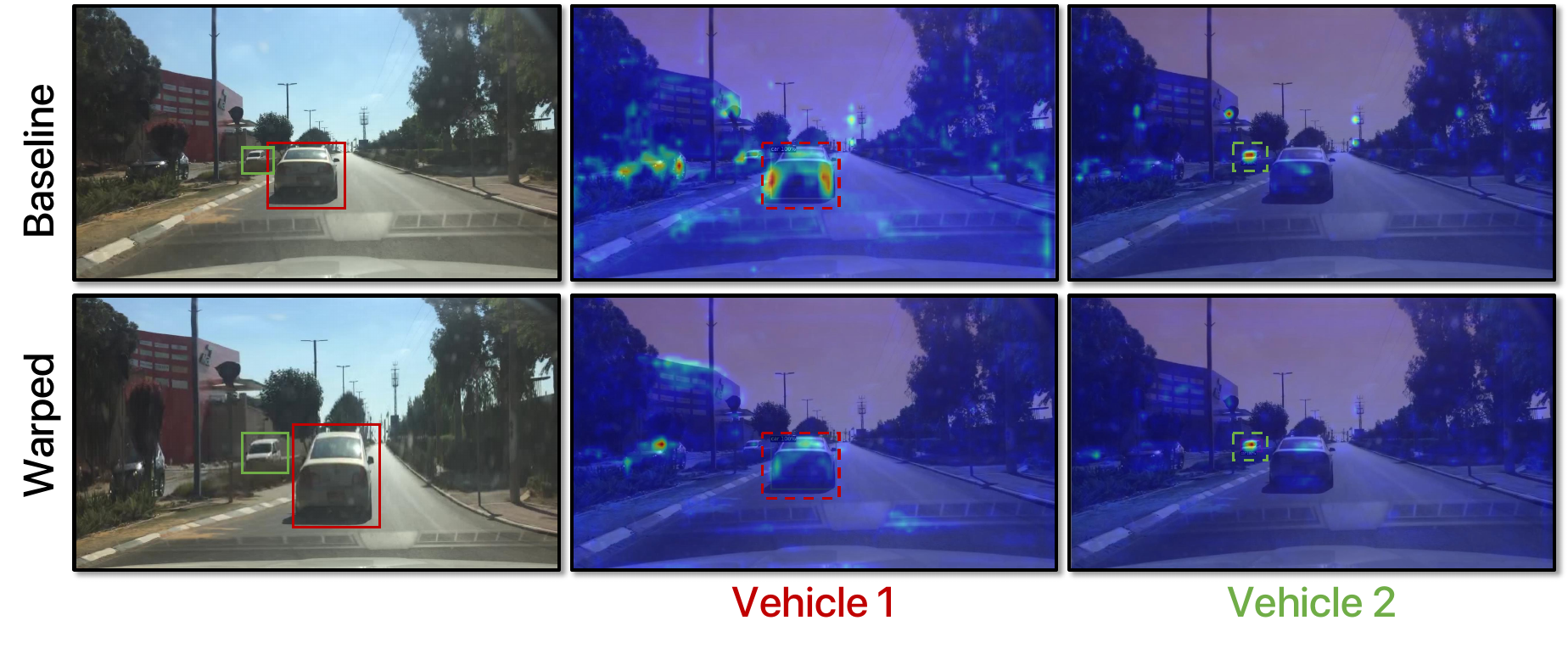} 
    \caption{\textbf{Additional Analysis: Grad-CAM~\cite{selvaraju2017grad} Visualizations for Multiple Objects.} We observe that learnt features show smaller Gradcam visualization contributions from background pixels compared to foreground object pixels when trained with instance-level warping. This is true for individual object instances, in this case, for Vehicle 1 and Vehicle 2. This suggests that our instance-level warping enhances the model's focus on foreground object pixels over background pixels. }
    \label{fig:grad_cam_explainer}
    \vspace{-0.1in}
\end{figure*}



\noindent \textbf{Grad-CAM Visualizations of the Learned Model:} In Section 5.2 of the main paper, we claimed that the learned backbone features obtained through our training method generalize better. To support this claim, we use GradCAM~\cite{selvaraju2017grad} visualizations to showcase ResNet features from detectors trained on BDD100K~\cite{yu2020bdd100k}, as shown in Figure~\ref{fig:grad_cam}. We observe that \textit{(a)} Heatmaps for models trained with 2PCNet show that with our warping, there is a strong focus on salient objects with minimal distracting activation, while without our warping, the focus is dispersed across the image, indicating a lack of precision. \textit{(b)} The choice of saliency guidance matters: our instance-level saliency guidance ensures a strong focus on salient objects, whereas warping guided by Geometric Prior~\cite{ghosh2023learned} results in less emphasis on target objects (e.g., cars) and an unnecessary focus on the background.

\noindent \textbf{Grad-CAM Visualizations for Multiple Objects:} Using the same backbone layers, Figure~\ref{fig:grad_cam_explainer} shows that Grad-CAM contributions from background pixels are smaller compared to those from foreground object pixels when our method is used during training. This demonstrates improved focus on foreground objects over the background when trained with our instance-level saliency guidance.

\begin{figure*}[t]
    \centering
     \includegraphics[width=0.7\textwidth]{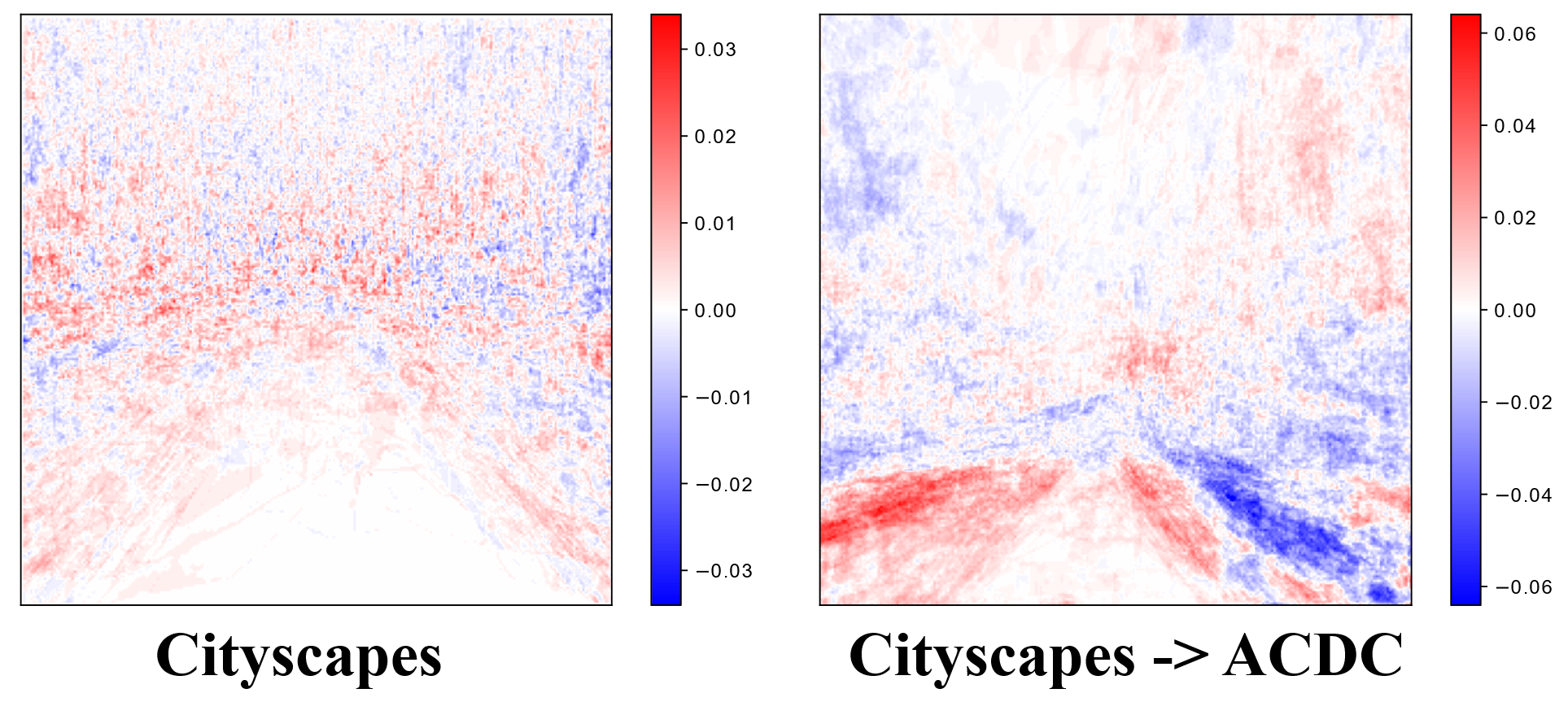}

    \caption{\textbf{Additional Analysis: Per-Pixel Accuracy Difference Visualization}. The visualization is done between baseline and ours. \textcolor{red}{Red} indicates our method is better, whereas \textcolor{blue}{blue} means the baseline is better. To improve the quality of visualization, we omit the comparison for `sky' and reshape the semantic segmentation maps to 256 $\times$ 256. We notice that our method outperforms the baseline across most pixels, except for the right-hand side sidewalk pixels in the ACDC dataset (\textcolor{blue}{blue} band) due to a significant disparity in width of the sidewalk (a background element) compared to the Cityscapes dataset.}
    \label{fig:err_vis}
\end{figure*}

\noindent \textbf{Per-Pixel Accuracy Difference Visualizations:} Figure~\ref{fig:err_vis} shows per-pixel accuracy difference visualizations. For both Cityscapes~\cite{cordts2016cityscapes} and ACDC~\cite{sakaridis2021acdc}, a noticeable predominance of red over blue is observed, indicating a clear advantage of our method over DAFormer~\cite{hoyer2022daformer} for semantic segmentation.

\begin{figure*}[t]
   \centering

\includegraphics[width=1\textwidth]{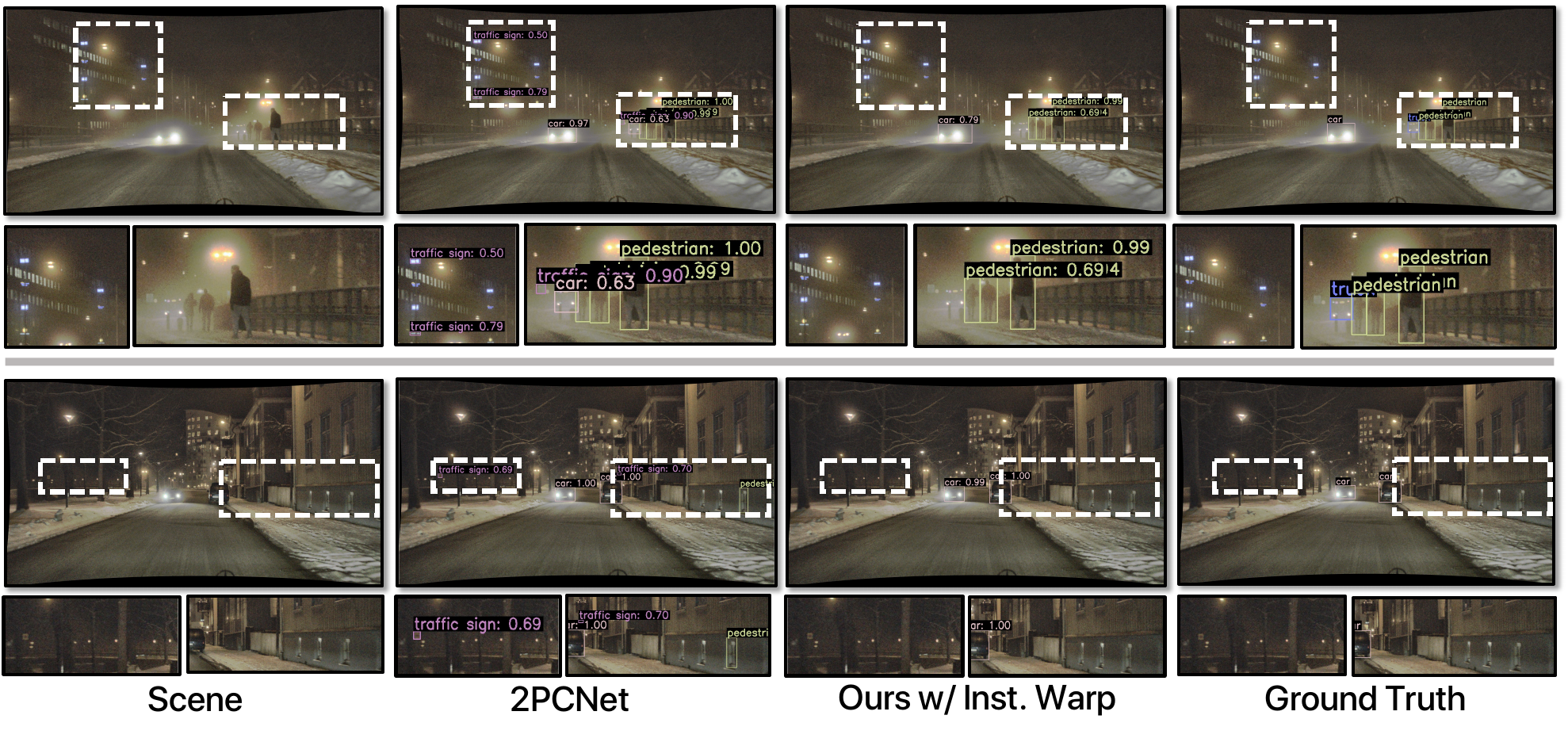}
   
   \caption{\textbf{Additional Analysis: BDD100K Clear $\rightarrow$ DENSE Foggy Object Detection.} Our method demonstrates superior object detection under foggy conditions. It accurately identifies streetlights and vehicles, which 2PCNet mislabels as traffic signs. Additionally, our method correctly ignores windows, which 2PCNet misclassifies as pedestrians.}
        \label{fig:bdd_dense_foggy}
\end{figure*}

\noindent
\textbf{BDD100K Clear $\rightarrow$ DENSE Foggy Object Detection:} Qualitative comparisons are shown in Figure~\ref{fig:bdd_dense_foggy}.  Our method demonstrates superior object detection under real foggy conditions by accurately identifying objects like streetlights and vehicles. In contrast, 2PCNet~\cite{kennerley20232pcnet} misidentifies windows as pedestrians, a mistake our approach avoids.
   
\begin{figure*}
    \centering

    \includegraphics[width=\textwidth]{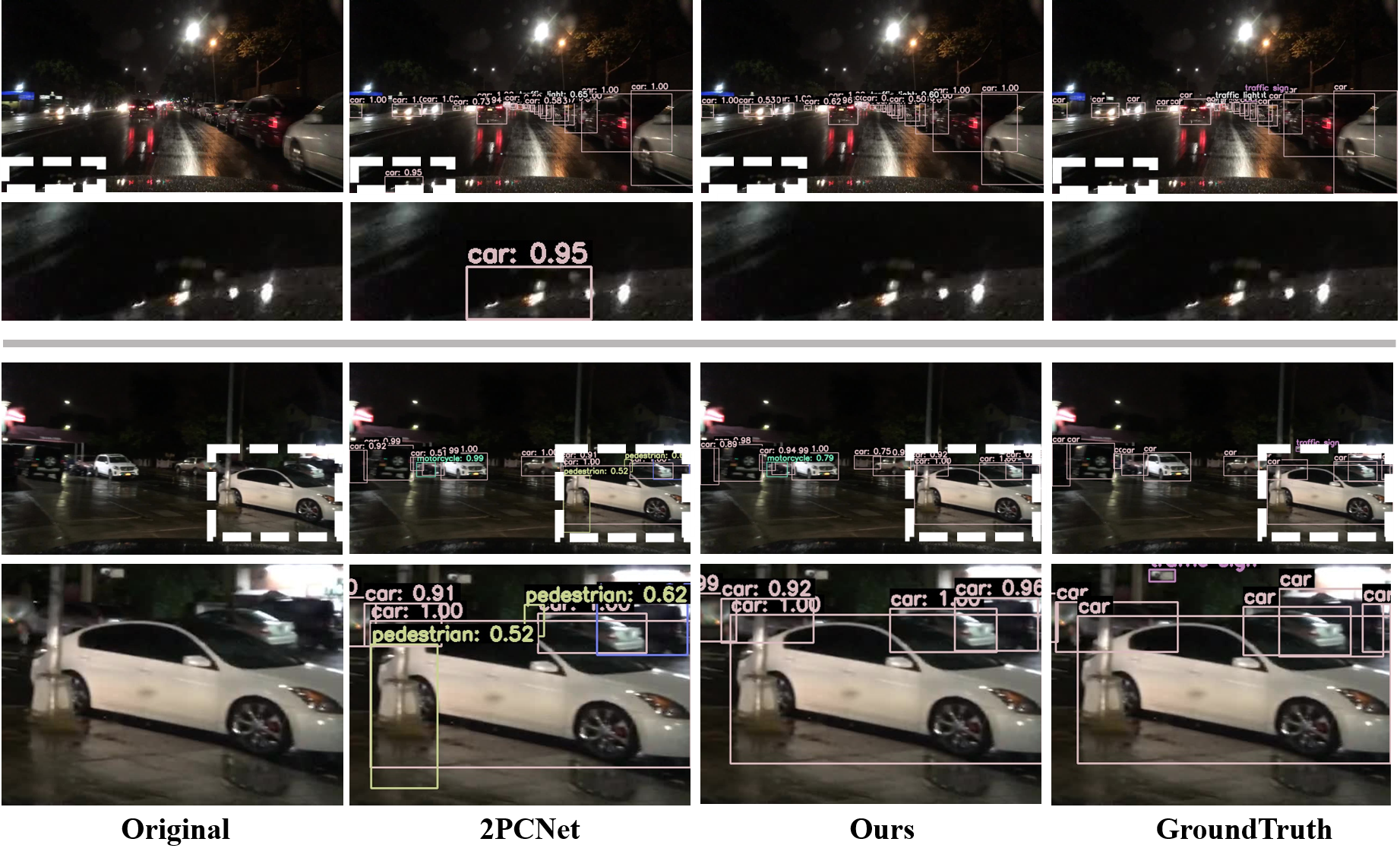}
    
    \caption{\textbf{Additional Analysis: BDD100K (Clear $\rightarrow$ Rainy) Object Detection.}  Our method demonstrates superior object detection under rainy conditions by accurately identifying vehicles and minimizing false positives, such as misidentified pedestrians and cars evident in 2PCNet~\cite{kennerley20232pcnet} predictions. }

    \label{fig:det_clear_rainy}
\end{figure*}

\noindent
\textbf{BDD100K (Clear $\rightarrow$ Rainy) Object Detection:} Qualitative comparisons are shown in Figure~\ref{fig:det_clear_rainy}. Our method demonstrates superior object detection under rainy conditions by accurately identifying vehicles and minimizing false positives, such as misidentified pedestrians and cars in the 2PCNet ~\cite{kennerley20232pcnet} predictions.

\begin{figure*}
    \centering
    \includegraphics[width=\textwidth]{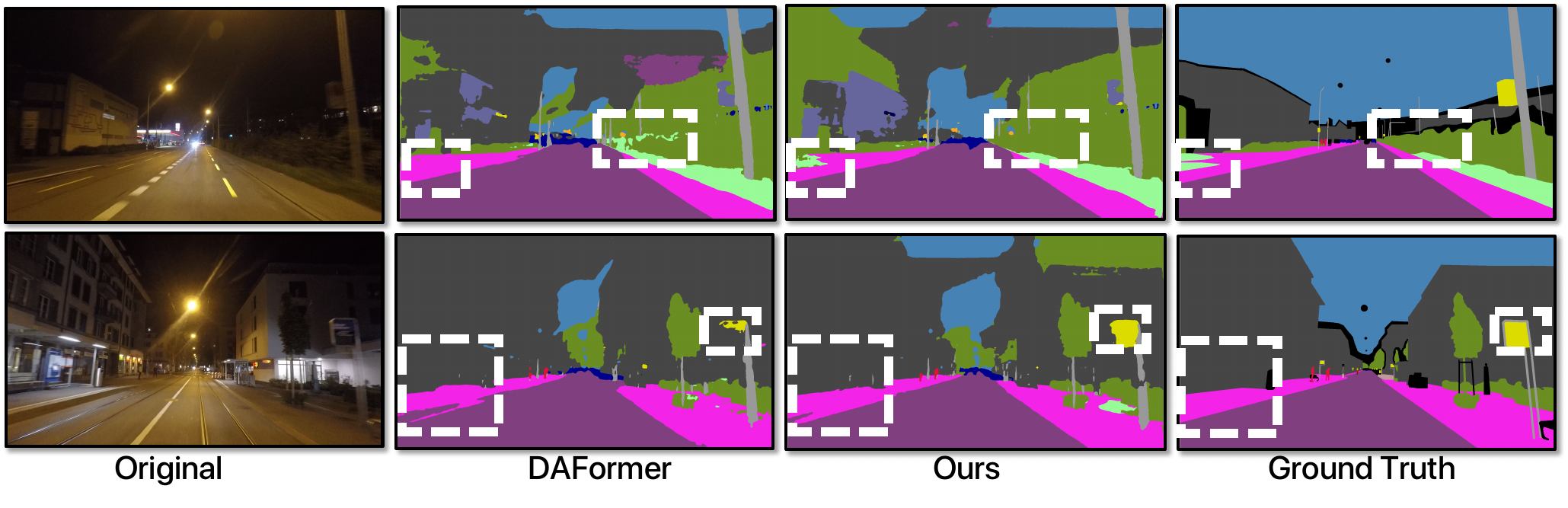}

    \caption{\textbf{Additional Analysis: Cityscapes $\rightarrow$ DarkZurich Semantic Segmentation.} Our method exhibits superior semantic segmentation in challenging low-light conditions. It produces segmentation outputs that closely align with the ground truth, particularly in accurately predicting road boundaries and consistently identifying urban elements such as sidewalks, terrain, and traffic signs.}
    \label{fig:seg_dz}
\end{figure*}

\noindent
\textbf{Cityscapes $\rightarrow$ DarkZurich Semantic Segmentation:}  Qualitative comparisons are shown in Figure~\ref{fig:seg_dz}. The proposed method produces segmentation outputs that more closely align with the ground truth, particularly in predicting road boundaries and consistently identifying urban elements such as sidewalks, terrain, and traffic signs. This indicates that our method has superior domain adaptation capabilities in challenging low-light conditions.

\clearpage
\section{Experimental Methodology}
\label{appendix:addl_details}

\noindent \textbf{Object Detection:} Following the recent and popular domain adaptation strategy 2PCNet~\cite{kennerley20232pcnet}, we use Faster R-CNN~\cite{ren2015faster} with ResNet-50~\cite{he2016deep}, adhering to their training hyperparameters and protocols. While 2PCNet~\cite{kennerley20232pcnet} focuses solely on Day-to-Night adaptation, our method addresses various adaptation scenarios. For scenarios other than Day-to-Night adaptation, we exclude the NightAug augmentation proposed by 2CPNet for both their method and ours.

\noindent\textbf{Semantic Segmentation:} We follow DAFormer~\cite{hoyer2022daformer} by employing the same SegFormer~\cite{xie2021segformer} head and MiT-B5~\cite{xie2021segformer} backbone, adhereing to their training hyperparameters, protocols, and seed for fair comparison. While Sim2Real Gap result presented in DAFormer~\cite{hoyer2022daformer} are not our focus, relevant results and discussion can be found in \pointToAppendix{\ref{appendix:sim2real}}{A}.

\noindent\textbf{Datasets:} We use BDD100K~\cite{yu2020bdd100k}, Cityscapes~\cite{cordts2016cityscapes}, DENSE~\cite{bijelic2020seeing},  ACDC~\cite{sakaridis2021acdc} \& DarkZurich~\cite{sakaridis2019guided}. A brief description is given below.

\noindent \textbf{BDD100K}~\cite{yu2020bdd100k} features 100,000 images with a resolution of 1280x720 for object detection and segmentation, covering various weather conditions and times of day, and includes annotations for 10 categories.

\noindent \textbf{Cityscapes}~\cite{cordts2016cityscapes} provides 5,000 images of urban road scenes at a resolution of 2048x1024 in clear weather for semantic segmentation, with annotations for 19 categories.

\noindent \textbf{DENSE}~\cite{bijelic2020seeing} provides 12,997 images at a resolution of 1920x1024, capturing diverse weather conditions such as heavy fog and heavy snow.

\noindent \textbf{ACDC}~\cite{sakaridis2021acdc} is designed for adverse conditions such as fog and snow, including 1,600 images at a resolution of 2048x1024 for segmentation across 19 categories.

\noindent \textbf{Dark Zurich}~\cite{sakaridis2019guided} is tailored for low-light conditions, offering 2,416 unlabeled nighttime images and 151 labeled twilight images for segmentation, all at a resolution of 1920x1080, with a focus on urban settings.

\clearpage
\section{Additional Technical Details}
\label{appendix:addl_tech}

\noindent
\textbf{Ground Truth Segmentation to Boxes:} Instance-level saliency guided image warping requires bounding boxes, which are not provided in some semantic segmentation benchmarks like GTA~\cite{richter2016playing}. To address this, we generate `from-seg' bounding boxes from ground truth semantic segmentation maps. Specifically, we first identify connected components representing individual instances of foreground categories, including traffic lights, traffic signs, persons, riders, cars, trucks, buses, trains, motorcycles, and bikes. For each `from-seg' instance, we then compute the bounding boxes by finding the minimum enclosing axis-aligned rectangle. These `from-seg' bounding boxes are finally used for instance-level saliency guidance in the same way as ground truth bounding boxes.

\clearpage







\end{document}